\documentclass[sigconf]{aamas} 

\usepackage{balance} 
\usepackage[utf8]{inputenc} 
\usepackage[T1]{fontenc}    
\usepackage{hyperref}       
\usepackage{url}            
\usepackage{booktabs}       
\usepackage{amsfonts}       
\usepackage{nicefrac}       
\usepackage{microtype}      
\usepackage{xcolor}         
\usepackage{subfigure}
\usepackage[bb=boondox]{mathalfa}
\usepackage{amsmath}
\usepackage{wrapfig}
\usepackage{multirow}
\usepackage{soul}
\usepackage{algorithmic}
\usepackage{enumitem}
\usepackage[ruled,linesnumbered]{algorithm2e}
\usepackage{caption}
\usepackage[figuresright]{rotating}
\usepackage{chngpage}

\newcommand{\hide}[1]{}

\newcommand{\name}{{ACE}}
\newcommand{\planner}{{MCP}}

\setcopyright{ifaamas}
\acmConference[AAMAS '23]{Proc.\@ of the 22nd International Conference
on Autonomous Agents and Multiagent Systems (AAMAS 2023)}{May 29 -- June 2, 2023}
{London, United Kingdom}{A.~Ricci, W.~Yeoh, N.~Agmon, B.~An (eds.)}
\copyrightyear{2023}
\acmYear{2023}
\acmDOI{}
\acmPrice{}
\acmISBN{}

\acmSubmissionID{767}

\title[AAMAS-2023 Formatting Instructions]{Asynchronous Multi-Agent Reinforcement Learning for \\Efficient Real-Time Multi-Robot Cooperative Exploration}

\author{Chao Yu$^{1,2*}$, Xinyi Yang$^{1*}$, Jiaxuan Gao$^{1*}$, Jiayu Chen$^{1,2}$, Yunfei Li$^{1}$, Jijia Liu$^{3}$, Yunfei Xiang$^{1}$,\\ Ruixin Huang$^{1}$, Huazhong Yang$^{1}$, Yi Wu$^{1,2,4\dag}$, Yu Wang$^{1\dag}$\\}
\affiliation{$^1$ Tsinghua University, $^2$ Shanghai Artificial Intelligence Laboratory, $^3$ Tongji University, $^4$ Shanghai Qi Zhi Institute \\
\country{$^*$ Equal Contribution $^\dag$ Corresponding Author \\}}
\email{zoeyuchao@gmail.com}

\begin{abstract}
We consider the problem of cooperative exploration where multiple robots need to cooperatively explore an unknown region as fast as possible. 
Multi-agent reinforcement learning (MARL) has recently become a trending paradigm for solving this challenge. 
However, existing MARL-based methods adopt \emph{action-making steps} as the metric for exploration efficiency by assuming all the agents are acting in a fully synchronous manner: i.e., \emph{every} single agent produces an action \emph{simultaneously} and \emph{every} single action is executed \emph{instantaneously} at each time step. 
Despite its mathematical simplicity, such a synchronous MARL formulation can be problematic for real-world robotic applications. It can be typical that different robots may take slightly different wall-clock times to accomplish an atomic action or even periodically get lost due to hardware issues. 
Simply waiting for every robot being ready for the next action can be particularly \emph{time-inefficient}.
Therefore, we propose an asynchronous MARL solution, \emph{Asynchronous Coordination Explorer (\name)}, to tackle this real-world challenge. We first extend a classical MARL algorithm, multi-agent PPO (MAPPO), to the asynchronous setting 
and additionally apply action-delay randomization to enforce the learned policy to generalize better to varying action delays in the real world. Moreover, each navigation agent is represented as a team-size-invariant CNN-based policy, which greatly benefits real-robot deployment by handling possible robot lost and allows bandwidth-efficient intra-agent communication through low-dimensional CNN features.
We first validate our approach in a grid-based scenario. Both simulation and real-robot results show that {\name} reduces over 10\% actual exploration time compared with classical approaches. We also apply our framework to a high-fidelity visual-based environment, Habitat, achieving $28\%$ improvement in exploration efficiency. 
\end{abstract}


\newcommand{\BibTeX}{\rm B\kern-.05em{\sc i\kern-.025em b}\kern-.08em\TeX}

\begin{document}


\pagestyle{fancy}
\fancyhead{}


\maketitle 


\section{Introduction}

Exploration is a fundamental task for building intelligent robot systems, which has been applied in many application domains, including rescue~\cite{rescue}, autonomous driving~\cite{autonomousdriving}, drone~\cite{drone}, and mobile robots~\cite{robots}. In this paper, we consider a multi-robot cooperative exploration task, where multiple homogeneous robots simultaneously explore an unknown spatial region in a cooperative fashion. 
Learning the optimal cooperative strategies can be challenging due to the existence of multiple robots. These robots must effectively distribute the exploration workload so that they can always navigate towards different spatial regions to avoid trajectory conflicts, which accordingly leads to a remarkably higher exploration efficiency than the single-robot setting. 

Multi-agent reinforcement learning (MARL) has been a trending approach to tackle this cooperative exploration challenge. RL-based methods directly learn neural policies end to end by interacting with a simulated environment for policy improvement. 
Compared with planning-based solutions~\cite{singleagent-RL2,frontier3,RRT} which require non-trivial implementation heuristics and expensive inference computation at execution time, RL-based methods~\cite{singleagent-RL2,singleagent-RL1} provide strong representation capabilities of complex strategies and negligible inference overhead once the policies are trained. 

Classical multi-agent RL algorithms typically adopt a synchronous algorithmic framework, i.e., all the agents are making actions at the same time, and all the actions will be executed immediately at each time step, leading to the next action-making step for future actions. This process is mathematically formulated as a decentralized Markov decision process, which is widely adopted in multi-agent RL literature.
Although such a mathematical framework is simple and elegant, it can be problematic for real-world multi-robot exploration tasks. 
For real robot systems, each actual action is never atomic and may take varying times to finish. These action delays can be more severe due to unexpected network communication traffic or hardware failure. Simply following the synchronous setting, i.e., waiting until every robot is ready before making new actions, can be particularly real-time inefficient. 
Therefore, an ideal RL framework for real-world use should be asynchronous, i.e., whenever an agent finishes action execution, it should immediately generate the next action, and the learned strategy should effectively enable such an asynchronous action-making process.

In this paper, we propose a novel asynchronous multi-agent RL-based solution, \emph{Asynchronous Coordination Explorer ({\name})}, to tackle the real-world multi-robot exploration task. We first extend a classical MARL algorithm, multi-agent PPO (MAPPO), to the asynchronous setting to effectively train the multi-agent exploration policy, and additionally leverage an action-delay randomization technique to enable better simulation to real-world generalization. 
Moreover, we design a communication-efficient Multi-tower-CNN-based Policy ({\planner}) for each agent. In {\planner}, a CNN module is applied to each agent's local information to extract features, and a fusion module combines each agent's features to produce an action. During execution time, efficient intra-agent communication can be achieved via directly exchanging low-dimensional features extracted by the weight-sharing CNN module. Another benefit of {\planner} is to tackle varying team sizes, which may occur when agents go offline in real-world applications.

We conduct experiments in a grid-based multi-room scenario both in simulation and our real-world multi-robot laboratory, where strategies learned by {\name} significantly outperform both classical planning-based methods and neural policies trained by \hide{classical }synchronous RL methods. 
In particular, \emph{{\name} reduces $10.07\%$ \textbf{real-world} exploration time} than the synchronous RL baseline and \emph{reduces $33.86\%$ \textbf{real-world} exploration time} than the fastest planning-based method with 2 Mecanum steering robots. Besides, we \hide{also }extend {\name} to a vision-based environment, Habitat, verifying the effectiveness of the asynchronous training mechanism when applied to more complicated environments. More demonstrations can be seen on our website: \url{https://sites.google.com/view/ace-aamas}.

\section{Related Work}

\subsection{Cooperative Exploration}
Multi-agent cooperative exploration is an important task for building intelligent mobile robot systems. There are a large number of works developing planning-based methods for this problem~\cite{sample3,RRT,yamauchi1997frontier}, but they typically rely on manually designed heuristics~\cite{multi-classical1,multi-classical2,frontier3} and are limited in expressiveness to learn more complex cooperation strategies. Another popular line of research is deep MARL-based methods which leverage the expressiveness power of neural networks to learn non-trivial cooperative exploration skills~\cite{multiagent-navigation1,multiagent-RL2,Wang*2020Influence-Based,epciclr2020,Yang2020CM3,DBLP:conf/aaai/WangYLHHHCFG20}. Note that most existing MARL-based methods assume synchronous action execution among all the agents or consider atomic actions, which we believe is due to the synchronous design of most simulated RL environments. However, such synchronous design does not reflect the real-world multi-agent systems, where agents take actions at different real times due to network delay and unexpected hardware take-downs. We propose an \emph{asynchronous} MARL exploration framework in this work to better match real-world applications. \cite{DBLP:journals/corr/OmidshafieiAAH15} considers an asynchronous decision-making mechanism for large-scale problems and proposes an improved Monte Carlo Search method to solve this problem. A concurrent work formulates a set of asynchronous multi-agent actor-critic methods that allow agents to directly optimize asynchronous policies~\cite{https://doi.org/10.48550/arxiv.2209.10113}, while we design an asynchronous MARL training framework combined with action-delay randomization. We also notice a recent trend of developing asynchronous simulation~\cite{jia2020fever,yao2021partially}, which we hope can further accelerate the advances in applying MARL methods to the real world.  

\subsection{Sim2real Transfer}
It is often challenging to directly deploy policies trained in simulation to the real world since there is always a mismatch between simulation and reality. Domain randomization is a simple but effective technique to fill the reality gap, which creates a variety of simulated environments with randomized properties such as physical dynamics~\cite{tobin2017domain,DBLP:journals/corr/abs-1710-06537} and visual appearances~\cite{DBLP:journals/corr/TobinFRSZA17,DBLP:journals/corr/abs-1808-00177}, and tries to train RL that can perform well among all of them. We also adopt the idea of domain randomization, and randomly delay the execution of each agent in simulation to model the uncertain delay between policy computation and actual action execution in the real world. The action delay technique is also adopted in other domains such as model-based RL~\cite{CHEN2021119} and reactive RL~\cite{10.3389/frobt.2018.00079}.

\subsection{Multi-Agent Communication}
Most works in MARL only consider perfect communication where agents can receive messages from all other agents~\cite{zhang2021multi,godoy2015adaptive,ding2018hierarchical}, but the requirements on communication bandwidth and transmission rate are costly. Recent works have begun to focus on learning efficient communication. \cite{NIPS2016_c7635bfd,NIPS2016_55b1927f} learn communication protocols in limited-bandwidth communication channels. Some works propose to learn which agent to communicate with using attention mechanisms~\cite{NEURIPS2018_6a8018b3,niu2021multi} or weight-based schedulers~\cite{kim2018learning}.
In our work, we focus on how to efficiently communicate with other agents with limited information to maintain perfect decision performance. We propose an extraction module to obtain essential features from high-dimensional information, which is much more efficient than delivering observation information directly.

\subsection{Size-Invariant Representation}
Attention mechanism~\cite{transformer} is widely used in RL policy representation to capture object-level information~\cite{duan2017one,wang2018non,graphnetwork,deepset} and represent relations~\cite{zambaldi2018relational,singleagent-RL2}. In MARL, attention-based policy can also be applied for generalization to an arbitrary number of input entities~\cite{epciclr2020,wang2020few,DBLP:journals/corr/abs-2011-08055,duan2017one,dgn,hama,wang2018non}. There are some other generalization settings in MARL, such as other-play~\cite{pmlr-v119-hu20a}, population-based training~\cite{osindero2017population,doi:10.1126/science.aau6249} and zero-shot team formation~\cite{baker2020emergent}. In addition, some training techniques are used to better deal with varying sizes of agents, such as evolutionary learning ~\cite{epciclr2020,czarnecki2018mix} and curriculum learning~\cite{chen2021variational}. In our work, we design a Multi-tower-CNN-based Policy based on an attention mechanism to tackle varying team sizes. Parameter-sharing is a commonly used paradigm for varying team sizes in MARL with homogeneous agents, which is also adopted in this work. It learns an identical policy for each agent and helps to reduce nonstationarity and improve training efficiency~\cite{DBLP:journals/corr/abs-1710-00336, DBLP:journals/corr/abs-2005-13625}.

\section{Preliminary}

\subsection{Task Setup} \label{task-setup}

We study the task of real-time multi-robot cooperative exploration, where a team of robots aims to explore an unknown environment exhaustively as fast as possible. Robots can transmit local information to each other to avoid duplicate exploration for better efficiency. 
The real-time multi-robot task is with an asynchronous nature, i.e., different robots do not take actions and receive the next-step observations at the same time. A real robot often requires non-fixed time to execute an action, and unexpected hardware failure can cause random delays. Besides, under the standard bi-level control setting in robot navigation~\cite{utility, yamauchi1997frontier, APF, RRT}, each action is a goal position to reach and typically requires varying number of atomic steps to accomplish, thus exacerbating the asynchronous issue.
The asynchronous execution is not considered by classical multi-agent reinforcement learning (MARL) works. It is typically assumed all agents take actions at the same \emph{action-making step} and do not take the action execution time into account. In the traditional MARL literature~\cite{surprising2021yu}, the task  is usually formulated as a decentralized partially observable Markov decision process (Dec-POMDP), which is unable to capture the asynchronous property in our setting. 


\subsection{Problem Formulation}
\begin{figure*}[ht!]
    \centering
    \vspace{-5mm}
    \includegraphics[width=0.8\linewidth]{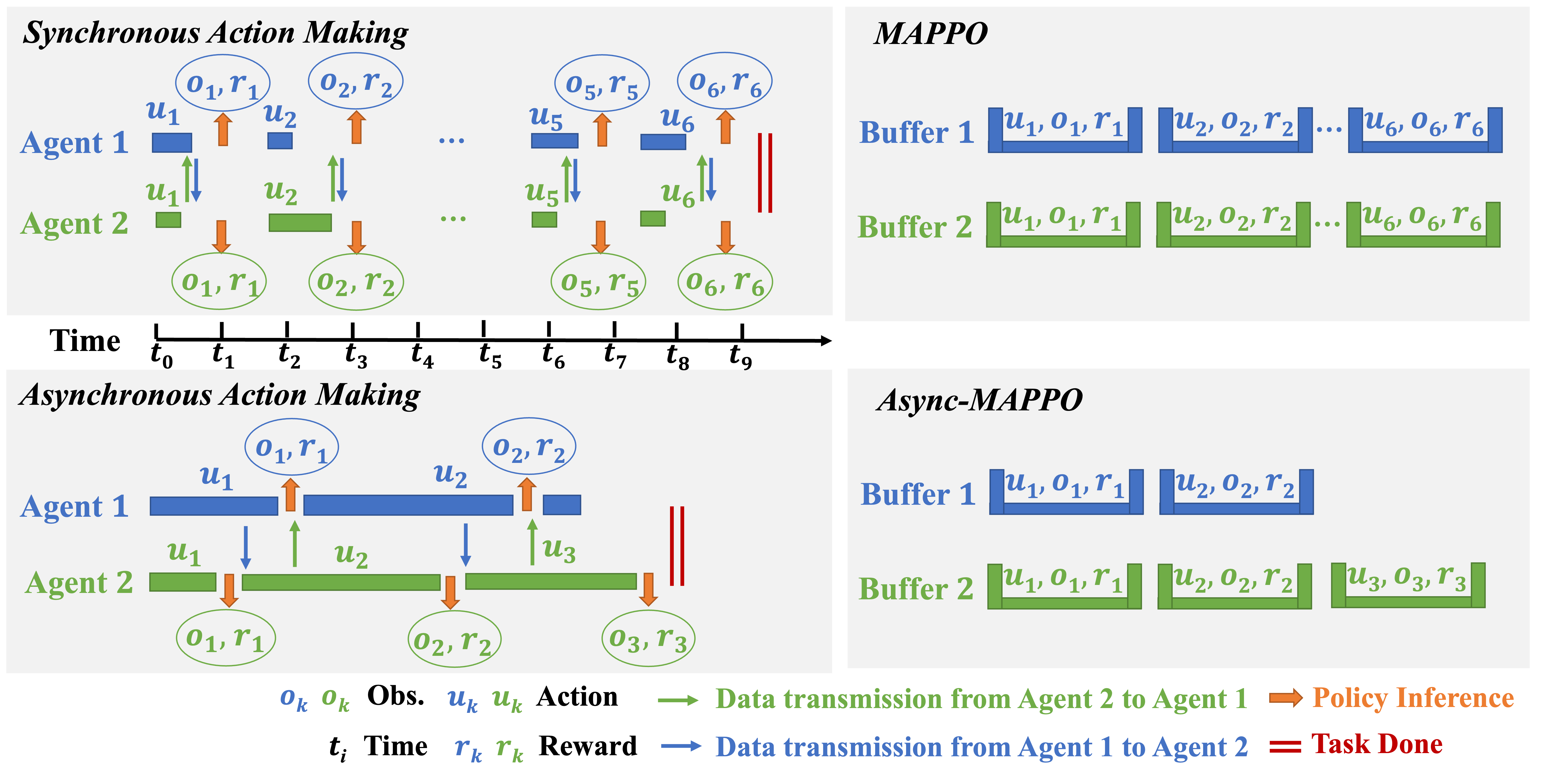}
	\centering 
	\vspace{-5mm}
	\caption{Comparison of \emph{asynchronous} and \emph{synchronous} action making.}
 \vspace{-3mm}
\label{fig:pipeline}
\end{figure*}
We model the asynchronous multi-agent cooperative exploration task as a decentralized partially observable Semi-Markov decision process (Dec-POSMDP)~\cite{DBLP:journals/corr/OmidshafieiAAH15} with shared rewards. We adopt a modular action execution scheme~\cite{ans,doi:10.1126/scirobotics.abg5810} which consists of bi-level actions for robust deployment in real-world robot systems. A macro action (MA), i.e., global goal, is generated in the action-making step. Several atomic actions, i.e., execution actions, are followed to perform under the guidance of the MA. 

To avoid notation ambiguity, we use $p^{(i)}$ to denote a parameter $p$ related to the i-th agent, and $\bar p=(p^{(1)},p^{(2)},\cdots, p^{(n)})$ to denote joint parameters for multiple agents thereafter. A Dec-POSMDP is defined by a set of elements $\langle D, \bar U, \bar B, P, \bar R^{\tau}\rangle$.
$D=\langle \bar S, \bar A, \bar \Omega, \bar O, \bar R, P, n, \gamma\rangle$
defines the decentralized partially observable Markov decision processes (Dec-POMDP), where $\bar S$ is the joint state space, $\bar A$ is joint atomic action space, $\bar \Omega$ is the observation space, $O^{(i)}(o^{(i)}|s,a^{(i)})$ denotes the observation probability function for agent $i$, $\bar R:\bar S\times \bar A\rightarrow \mathbb R$ is the joint reward function, $n$ is the number of agents, $P$ is the state transition probability.
$\bar U$ is joint macro-action space. 
A macro action $u^{(i)}$ is a high-level policy that can generate a sequence of atomic actions $a_t\sim u^{(i)}(H_t^{(i)})$ for any $t$ when $u^{(i)}$ is activated, 
where $H_t^{(i)}$ is the individual action-observation history till $t$.
$\bar B$ denotes the stop condition of MA and $B^{(i)}(u^{(i)})$ is represented as a set of action-observation histories of an agent $i$. If $H_t^{(i)}\in B^{(i)}(u^{(i)}_t)$ holds, $u_{t}^{(i)}$ terminates and the agent generates a new MA.
$\bar R^{\tau}$ is the macro joint reward function: 
$\bar R^{\tau}(\bar s, \bar u) = \mathbb{E}\left[\sum_{t=0}^{\bar \tau_{end}}\gamma^t \bar R(\bar s_t, \bar a_t)|\bar a_t\sim \bar u(\bar H_t)\right]$ where $\bar{\tau}_{end}=\min_{t}\{t:H_t^{(i)}\in B^{(i)}(u^{(i)})\}$.
        
The solution of a Dec-POSMDP is a joint high-level decentralized policy $\bar \phi=(\phi^{(1)},\cdots, \phi^{(n)})$ where each $\phi^{(i)}$ produces an MA $\phi^{(i)}(H_t^{(i)})\in U^{(i)}$ given individual action-observation history $H_t^{(i)}$.
In the beginning of an episode, an initial MA is computed as: $u_{t_0}^{(i)}=\phi^{(i)}(H_{t_0}^{(i)})$. At action-making step $k > 0$, the agent generates a new MA $u_{t_k}^{(i)}=\phi^{(i)}(H_{t_k}^{(i)})$ if the stop condition is met, i.e. $H_{t_k}^{(i)}\in B^{(i)}(u_{t_{k-1}}^{(i)})$. Otherwise, the agent continues to use the previous MA: $u_{t_k}^{(i)}=u_{t_{k-1}}^{(i)}$. In the time range $[t_k, t_{k+1})$, the agent interacts with the environment with atomic actions sampled from MA: $a_t^{(i)}\sim u^{(i)}(H_t^{(i)})$.
Finally, the goal of Dec-POSMDP is to maximize the accumulative discounted reward: $\mathbb E\left[ \sum_{k=0}^{\infty}\gamma^{t_k}\bar R^{\tau}(\bar s_{t_k}, \bar u_{t_k})|\bar\phi, \bar s_0 \right]$ where $t_0=0$ and $t_k=\min_{t}\{t>t_{k-1}:H_t^{(i)}\in B^{(i)}(u^{(i)}_{t_{k-1}})\}$ for $k\ge 1$.
A more detailed definition can be found in \cite{DBLP:journals/corr/OmidshafieiAAH15}.

In our \emph{asynchronous} setting, $t$ is the real time, not the discrete time step as in common \emph{synchronous} RL. Our setting is more time-efficient and robust to hardware faults. Take a 2-agent case as an example (see Fig.~\ref{fig:pipeline}), in the synchronous setting, the agents can only transmit data (blue and green arrows) and perform policy inference (orange arrow) after both of them have finished the previous action execution.
The system execution speed is bottle-necked by the agent with the longest execution time. Worse still, the whole system will get stuck if one agent goes offline unexpectedly. 
By contrast, agents take actions in a distributed manner in an asynchronous setting. Each agent can request data from other agents and conduct policy inference immediately after it finishes its own action execution.
This asynchronous setting is more time-efficient for multi-agent exploration tasks, and will not be blocked by dynamic changes such as agents going offline. 

\subsection{Connection to Conventional MARL}

In the conventional MARL literature~\cite{surprising2021yu}, the problem formulation is typically under decentralized partially observable Markov decision process (Dec-POMDP), which assumes synchronized actions. In this work, we also focus on the multi-agent setting and assume a shared reward function and dynamic transitions. However, different from synchronous MARL which assumes all agents execute actions simultaneously, we consider the asynchronous nature in the practical multi-robot scenarios. 

We will adapt a popular MARL algorithm, Multi-Agent Proximal Policy Optimization (MAPPO)~\cite{surprising2021yu}, from the conventional setting to our asynchronous setting.  
Conventional MAPPO follows the Centralized Training and Decentralized Execution (CTDE) paradigm, in which agents make decisions with individual observations and update the joint policy with global information in a centralized manner. Under the framework of Dec-POMDP, MAPPO requires all agents taking actions synchronously at each discrete time step, and the state transits according to actions from all agents: $s_t\sim P(\cdot|s_{t-1},\bar a_{t-1})$. It aims to find a joint policy $\bar\pi$ that maximizes the accumulated discounted reward $\mathbb E\left[\sum_{t=0}^{\infty} \gamma^t \bar R(s_t,\bar a_t)|a_t^{(i)}\sim \pi^{(i)}(H_t^{(i)})\right]$. 
Different from MAPPO, Async-MAPPO is designed for the asynchronous setting, where there are no centralized environment steps.


\section{Methodology}

To better model the asynchronous nature of real-world multi-agent exploration problems, we present Asynchronous Coordination Explorer ({\name}). {\name} consists of 3 major components: (1) Async-MAPPO for MARL training, (2) action-delay randomization for zero-shot generalization in the real world, and (3) multi-tower-CNN-based policy representation for efficient communication. 

\subsection{Async-MAPPO}\label{sec:asyncmappo}

We extend an on-policy MARL algorithm MAPPO~\cite{mappo} to our asynchronous setting, which we call Async-MAPPO. The pseudo-code of Async-MAPPO is shown in Algo.~\ref{algo:async-mappo}. Compared with the setting of MAPPO, both policy execution and data collection are not necessarily time-aligned among different agents, and we implement the asynchronous action-making and replay buffer as follows.


\begin{enumerate}[nolistsep,leftmargin=*]
    \item[$\bullet$] We design a bi-level execution scheme. In {\name}, agents perform atomic actions under the guidance of global goals (macro actions). Instead of receiving the reward, local observation, and states immediately after executing an atomic action, Async-MAPPO accumulates the reward between action-making steps and only takes observation and states at each macro action.
    \item[$\bullet$] We implement asynchronous buffer insertion, in contrast to the synchronous scheme in original MAPPO as shown in Fig.~\ref{fig:pipeline}. The original MAPPO assumes synchronous execution of all the agents; in each time step, all the agents take actions simultaneously, and the trainer waits for all the new transitions before inserting them into a centralized data buffer for RL training. In Async-MAPPO, different agents may not take actions at the same time (some agents may even get stuck and cannot return new observations at all), which makes it infeasible for the trainer to collect transitions in the original synchronous manner. Therefore, we allow each agent to store its own transition data in a separate cache and periodically push the cached data to the centralized data buffer. We can then run the standard MAPPO training algorithm over this buffer.
\end{enumerate}

\begin{algorithm}
\caption{Async-MAPPO}
\label{algo:async-mappo}
Initialize the  policy $\pi$\;
\While{$step\le step_{max}$}{
    set data buffer $D=\{\}$\;
    \For{$i=1$ to $batch\_size$}{
        Reset the environment\;
        Create $N$ empty caches $C=[[], \dots, []]$\;
        \For{$t=1$ to $T$}{
            \For{all agents $i=1$ to $N$}{
                \If{agent $i$ replans macro action}{
                    $b \gets$ agent $i$'s $b$-th macro actions\;
                    $s_{b}^{(i)}\leftarrow State, o_{b}^{(i)}\leftarrow Observation$\; 
                    $C_i+=[s_{b-1}^{(i)},o_{b-1}^{(i)},u_{b-1}^{(i)},\hat R_b^{\tau(i)},s_{b}^{(i)},o_{b}^{(i)}]$\;
                    $p_{b}^{(i)}=\pi(o_b^{(i)})$\;
                    Update macro action $u_b^{(i)}\sim p_{b}^{(i)}$\;
                }
                Execute atomic action $a_t^{(i)}\sim u_b^{(i)}$\;
            }
        }
        Compute reward-to-go and insert data into $D$\;
    }
    Update $\pi$ on MAPPO loss\;
}
\end{algorithm}

\subsection{Action-Delay Randomization}
\label{action-delay randomization}

When training in traditional simulators, agents can always take execution steps synchronously without considering different action execution costs. Moreover, real-world action delays such as hardware failure and network blocking are not simulated. These problems cause a large gap for deploying trained agents from simulation to reality. To reduce this gap, we apply action-delay randomization during simulation\hide{ to simulate the real-world challenge of action delay}. In the end of each action-making step, we force each agent to wait for a random period from $3$ to $5$ execution steps in grid-based environments, and from $10$ to $15$ execution steps in Habitat before querying the next macro action.
\begin{figure*}[ht!]
    \centering
    \vspace{-5mm}
\includegraphics[width=0.8\textwidth]{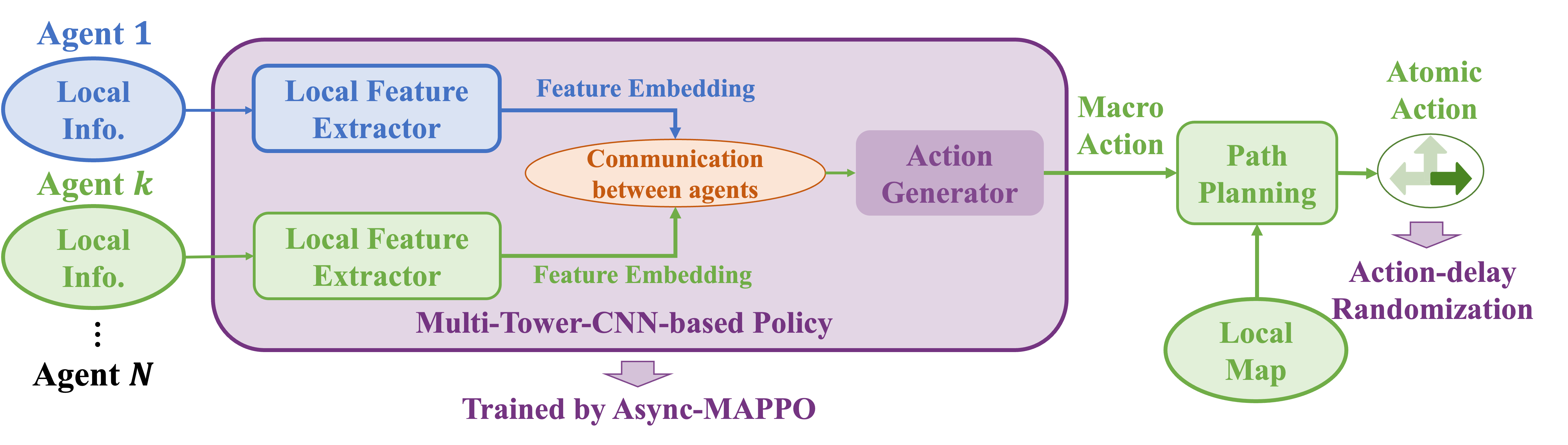}
    \vspace{-5mm}
    \caption{Overview of Asynchronous Coordination Explorer ({\name}).}
    \vspace{-3mm}
    \label{fig:overview}
\end{figure*}

\subsection{Multi-Tower-CNN-Based Policy}

\begin{figure}[h]
    \centering
    \vspace{-2mm}
    \includegraphics[width=0.9\linewidth]{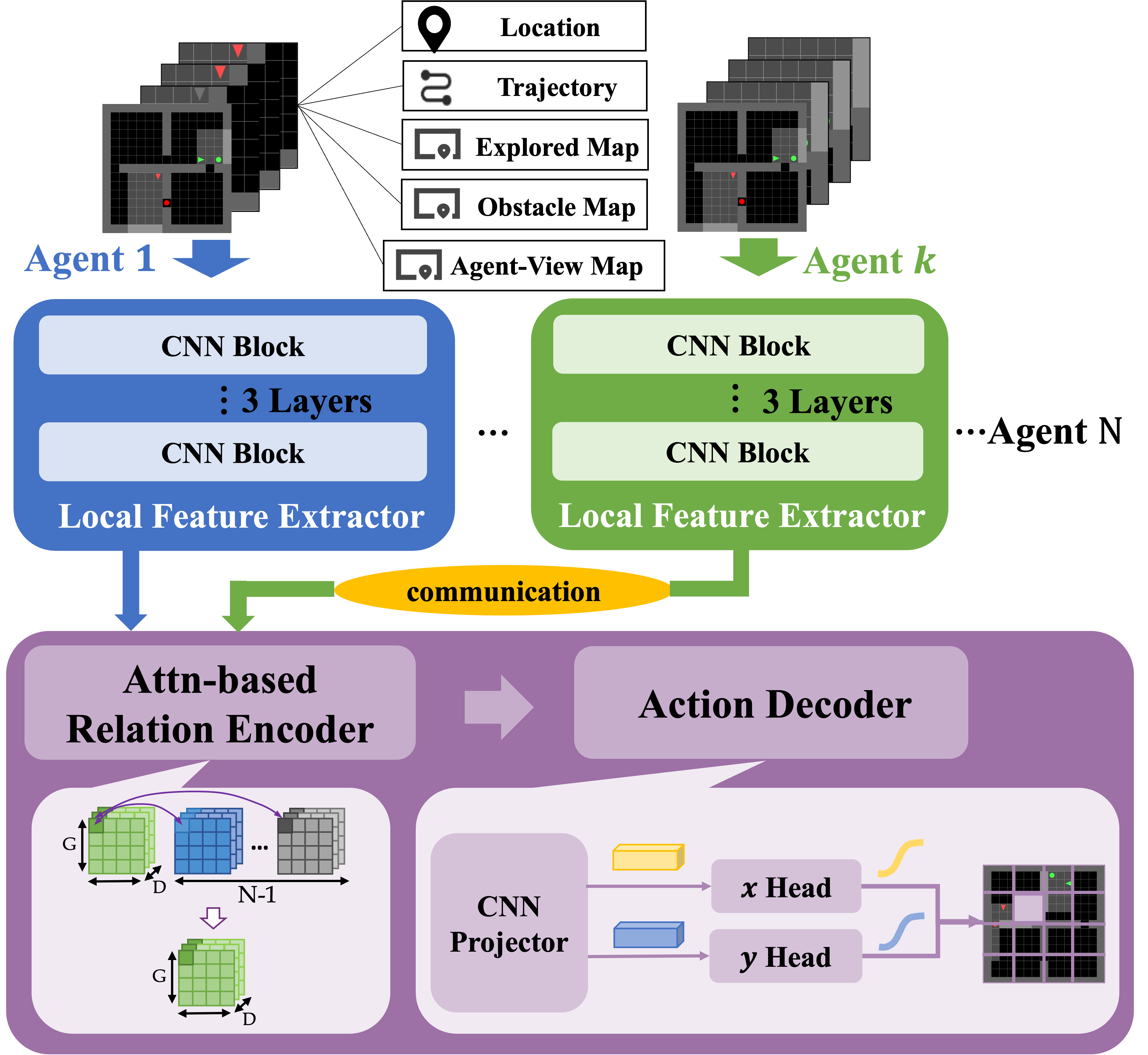}
    \centering 
    \vspace{-2mm}
    \caption{Workflow of \emph{Multi-tower-CNN-based Policy} (\planner), including a CNN-based local feature extractor, a relation encoder, and an action decoder.}
    \label{fig:ace}
    \vspace{-3mm}
\end{figure}

The Multi-tower-CNN-based Policy ({\planner}) is utilized to generate macro actions, i.e., global goals in {\name}. As illustrated in Fig.~\ref{fig:ace}, {\planner} consists of 3 parts, i.e., a CNN-based local feature extractor, an attention-based relation encoder, and an action decoder. 

The local feature extractor is a weight-sharing 3-layer CNN and can extract a $G\times G \times 4$ feature embedding from each agent's $S \times S\times 7$ local information, which includes one obstacle channel, one explored region channel, one-hot location channel, one trajectory channel to represent the history trace, and three agent-view channels of the agent's local observation.

The agents transmit extracted feature embedding instead of the raw local information, which greatly reduces communication traffic by $1-\frac{G\times G\times 4}{S\times S\times 7} = 1-\frac4{7\alpha^2}$ times where $\alpha=S/G$. For example, 
we adopt $G=5$ in grid-based environments, thus the communication traffic reduces $\sim97\%$ in $S=25$ maps and $\sim93\%$ in $S=15$ maps.

The relation encoder aims to aggregate the extracted feature maps from different agents to better capture the intra-agent interactions.
In team-based exploration, an agent should not only spot undiscovered areas but also inter-teammates' movement for better scheduling among agents. 
We adopt a simplified Transformer~\cite{transformer} block as the team-size-invariant relation encoder. 
Inspired by the vision transformer model~\cite{attention1}, we apply multi-head cross-attention~\cite{attention2} to derive a single team-size-invariant representation of size $G \times G \times 4$, as shown in Fig.~\ref{fig:ace}. 

Finally, the action decoder predicts the agent's policy from the aggregated representation as a multi-variable Categorical distribution to select a grid cell $g$ from a plane as the global goal $(u_x,u_y)$. Note that in Habitat, in order to produce accurate global goals, we adopt a spatial action space with three separate action heads, i.e., two discrete region heads for choosing a grid cell $g$, which are the same as grid-based environments, and two additional continuous point heads for outputting a coordinate $(\Delta_x,\Delta_y)$, indicating the relative position of the global goal within the selected region $g$. Details of {\planner} in Habitat can be found in Appendix A.2.

\subsection{Overall Architecture}

As shown in Fig.~\ref{fig:overview}, each agent observes the local information and requests the latest feature embedding from other agents, which is output by the weight-sharing local feature extractor, at each action-making step. That is, agents only need to transfer the low-dimensional feature embedding, instead of the entire local information. The multi-tower-CNN-based policy, which is trained by Async-MAPPO, generates the next macro action, i.e., global goal, at each action-making step, and the agent performs path planning on the local map according to the global goal, outputting the atomic action at each time step. Note that agents could go offline in multi-agent tasks due to unexpected network communication traffic or hardware failure.

\begin{figure*}[ht]
    \centering
    \vspace{-3mm}
    \includegraphics[width=0.8\textwidth]{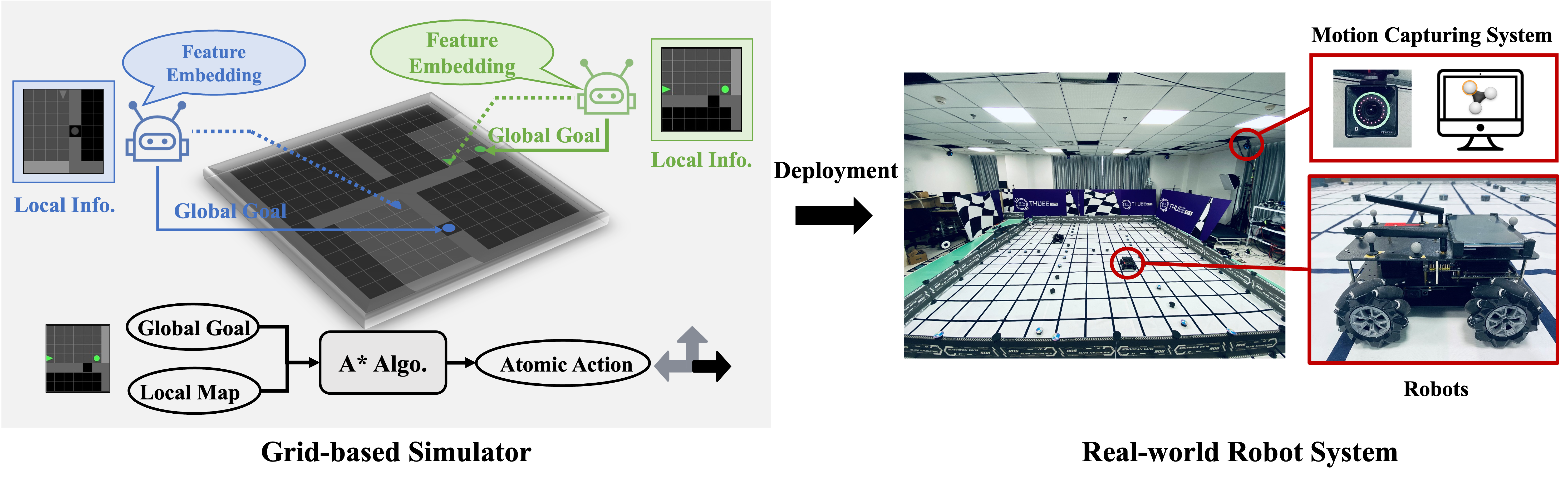}
    \vspace{-5mm}
    \caption{The illustration of the grid-based simulator and real-world robot system.}
    \vspace{-3mm}
    \label{fig:real}
\end{figure*}

\section{Environment Details}


Here we give details of the environments we adopted in this work, including the environment setting, the observation space and the action space of {\name}, as well as the designed reward function.
\subsection{Environment Setting}\label{deploy}

\textbf{Grid-based scenario:} As shown in Fig.~\ref{fig:real}, we implement a multi-agent exploration task based on the GridWorld simulator~\cite{gym_minigrid}, which was originally designed for synchronous settings. We consider two different map sizes, which are $15 \times 15$ with $4 \sim 9$ random rooms and $25 \times 25$ with $4 \sim 25$ random rooms. All the agents are uniform randomly spread over the map in the beginning. The local information of each robot is fed to the RL-trained policy or planning-based methods to generate a global goal and $A^{\star}$ algorithm is utilized to plan 5 atomic actions on the local map to follow the global goal.

We also set up a $15 \times 15$ real-world grid map which is the same as the grid-based simulation, and each grid is 0.31m long, as shown in Fig.~\ref{fig:real}. Our robots are equipped with Mecanum steering and an NVIDIA Jetson Nano processor. The locations and poses of robots are tracked by OptiTrack cameras and the Motive motion capture software. After training a policy in the grid-based simulator under $15 \times 15$ map with random rooms, we directly deploy it to the real-world robot system. Each real robot executes in a distributed and asynchronous manner. The robot adopts a request-send mechanism to obtain the newest feature embedding of other agents through ROS topic upon finishing all atomic actions.

\textbf{Habitat:}
We adopt map data from the Gibson dataset~\cite{dataset} while the visual signals and dynamics are simulated by Habitat~\cite{habitat}. We follow the same environment configuration in~\cite{maans} and use a pre-trained neural SLAM model to predict the robot pose and the local map. Full details of Habitat can be found in Appendix A. We also make sure that the birthplaces of agents are set to be close enough, i.e., agents are randomly scattered in a circle with a radius of 1 meter, so that the exploration task would be sufficiently challenging for learning. 

\subsection{Observation Space}
The input of RL-trained {\planner} is an $S\times S$ image with 7 channels, where $S$ is the max size of the map. The channels represent obstacles, the explored mask, the agent location, the trajectory, and three $H\times W$ agent-view. Note that each agent only maintains its locally observed information, which is memory and communication-efficient for real-world deployment. 

\subsection{Action Space}
The overall exploration framework is hierarchical, with a global goal (macro action) followed by several atomic actions towards the goal. The action of the policy is to generate a global goal $(u_x,u_y)$ chosen in the map, representing a discrete grid in grid-based environments or a continuous location in Habitat~\cite{habitat}. The available atomic actions are moving forward, turning left, and turning right provided by the simulator.
\vspace{-2mm}
\subsection{Reward Function}
The team-based reward function is the sum of the coverage reward, success reward, and overlap penalty.
Let $Ratio^t$ denote the total coverage ratio at time $t$, $Exp_a^t$ be the explored map by agent $a$ and $Exp^t$ denote the merged explored map by all agents. Both $Exp^t$ and $Exp^t_a$ are sets of explored areas. The reward terms are defined as follows. 

\begin{itemize}
    \item \textbf{Coverage Reward:} It is proportional to the size of the newly discovered region by the team $Exp^t\backslash Exp^{t-1}$. 
    \item \textbf{Success Reward:} Agent $a$ gets a success reward of $Ratio^t$ when $C\%$ coverage ratio is reached, which $C=98$ in the grid-like simulator and $C=90$ in Habitat\footnote{Maps in Habitat are harder than in the grid-based simulator, leading to differences in the success rate threshold.}.
    \item \textbf{Overlap Penalty:} The overlap penalty $r_{overlap}$ is designed to penalize repetitive exploration and encourage cooperation with others. It is defined as 
    \begin{equation*}
        \begin{array}{l}r_{overlap}=\left\{\begin{array}{l}
        \begin{array}{l}-A_{overlap}\times0.01,\;Ratio^t\;<0.9\\0,\;Ratio^t\ge0.9\end{array}\end{array} \right.
        \\
        \end{array},
    \end{equation*}
    where $A_{overlap}$ is the increment of the overlapped explored area between agent $a$ and other agents. The overlapped area between agent $a$ and agent $w$ is $Overlap^t_{a,w}=Exp^t_a\cap Exp^t_w$, and
    $A_{overlap} = \sum_ {w\in\{1,\cdots,n\}\backslash\{a\}} Overlap^t_{a,w}\backslash Overlap^{t-1}_{a,w}$.
\end{itemize}

\begin{table*}[ht!]
\vspace{-5mm}
\centering
\begin{tabular}{cccccc|cccc} 
\toprule
\multirow{2}{*}{Map Size} & \multirow{2}{*}{Methods} & \multicolumn{4}{c|}{Synchronous Action Making}         & \multicolumn{4}{c}{Asynchronous Action Making}                                        \\ 
\cmidrule{3-10}
                         &                         & Time $\downarrow$  & Overlap $\downarrow$ & Coverage $\uparrow$ & ACS $\uparrow$ & Time   $\downarrow$ & Overlap $\downarrow$ & Coverage $\uparrow$ & ACS $\uparrow$  \\ 
\midrule
\multirow{6}{*}{$15 \times15$}                     & Utility                  &      40.81(0.94)          & 0.45(0.02)            & \textbf{1.00(0.00)}           & 88.80(0.08)           & 35.75(0.99)          & 0.42(0.01)            & \textbf{1.00(0.00)}           & 90.01(0.14)       \\ 
\cmidrule{2-10}
                                           & Nearest                  &       25.44(0.53)         & 0.17(0.01)          & \textbf{1.00(0.00)}          & 91.60(0.17)      & 22.59(0.34)         & 0.17(0.01)           & \textbf{1.00(0.00)}         & 92.47(0.17)     \\ 
\cmidrule{2-10}
                                           & RRT                            & 28.86(0.99)         & 0.18(0.01)          & \textbf{1.00(0.00)}          & 91.46(0.03)         &25.85(0.36)        & 0.19(0.01)         & \textbf{1.00(0.00)}        & 92.36(0.08)      \\ 
\cmidrule{2-10}
                                           & APF                           & 24.95(0.76)         & 0.17(0.01)          & \textbf{1.00(0.00)}          & 91.57(0.39)          & 21.56(0.43)          & 0.17(0.01)          & \textbf{1.00(0.00)}        & 92.52(0.37)     \\ 
\cmidrule{2-10}
                                           & Voronoi                  & 48.57(3.64)        & 0.33(0.01)          & \textbf{1.00(0.00)}          & 86.43(0.33)          & 42.94(4.05)          & 0.20(0.01)          & \textbf{1.00(0.00)}        & 88.16(0.57)    \\ 
\cmidrule{2-10}
                                           & MAPPO                      & 24.75(0.45)   &  0.08(0.02)    &           \textbf{1.00(0.00)}  &          92.39(0.19)     &   21.92(0.90)             &  0.09(0.01)        &  \textbf{1.00(0.00)}     &  93.18(0.17)          \\ 
\cmidrule{2-10}
                                           & {\name}                    &   \textbf{21.76(0.79)} &  \textbf{ 0.07(0.00)}  &  \textbf{1.00(0.00) } & \textbf{92.54(0.21) } & \textbf{18.66(0.79)} & \textbf{0.07(0.01)}     &\textbf{1.00(0.00)}  &\textbf{93.39(0.14)}              \\ 
\midrule
\multirow{6}{*}{$25 \times25$}                     & Utility                       & 189.38(0.93)         &0.46(0.05)           & 0.93(0.01)          & 139.31(1.56)        & 183.71(1.59)        & 0.50(0.04)           & 0.95(0.01)          & 144.64(1.42)     \\ 
\cmidrule{2-10}
                                           & Nearest                       & 113.48(1.73)         & 0.24(0.01)          & \textbf{1.00(0.00)}          & 161.40(0.63)        & 99.52(2.00)         & 0.24(0.01)           & \textbf{1.00(0.00)}          & 166.53(0.91)      \\ 
\cmidrule{2-10}
                                           & RRT                          & 120.15(2.29)         & 0.20(0.01)           & \textbf{1.00(0.00)}          & 164.35(0.64)       & 105.64(1.69)         & 0.21(0.01)       & \textbf{1.00(0.00)}        & 168.33(0.27)      \\ 
\cmidrule{2-10}
                                           & APF                           & 101.41(0.78)         & 0.23(0.01)           & \textbf{1.00(0.00)}         & 162.46(0.64)       & 90.00(1.48)        & 0.23(0.01)           & \textbf{1.00(0.00)}          & 166.82(0.81)      \\ 
\cmidrule{2-10}
                                           & Voronoi                    & 131.65(0.41)       & 0.23(0.01)          & \textbf{1.00(0.00)}          & 160.92(0.33)          & 117.14(0.13)        & 0.20(0.01)          & \textbf{1.00(0.00)}        & 165.35(0.27)    \\ 
                                          
\cmidrule{2-10}
                                           & MAPPO                     &  90.15(1.08)         &    0.08(0.01)       &\textbf{1.00(0.00)}           &168.54(0.58)      &        82.55(2.70)            &  0.09(0.01)           & \textbf{1.00(0.00)}    &     171.72(0.27)            \\ 
\cmidrule{2-10}
                                           & {\name}             &  \textbf{83.34(0.44)} &  \textbf{0.06(0.00)}&   \textbf{ 1.00(0.00)}&   \textbf{170.03(0.49)} &  \textbf{74.36(2.93)} &  \textbf{0.06(0.00) } &\textbf{1.00(0.00)}
                   &\textbf{ 173.16(0.72)  }         \\
\bottomrule
\end{tabular}
\caption{Performances of different methods under 2-agent synchronous and asynchronous settings in the grid-based simulator.}
\label{tab: sync-async}
\vspace{-5mm}
\end{table*}
\begin{table*}[ht!]
\centering
\begin{tabular}{crccccc} 
\toprule
\multicolumn{1}{c}{\begin{tabular}[c]{@{}c@{}}\\\end{tabular}} &   Metrics       & Utility    & Nearest    & RRT        & APF        & {\name} \\ 
\midrule
\multirow{5}{*}{3 $\Rightarrow$ 2}                                           & Time $\downarrow$    & 139.33(1.79) &  76.53(1.16) &  81.86(1.14)&  74.80(2.11) &     \textbf{67.50(1.42)} \\
\cmidrule{2-7}
                                                               & Overlap $\downarrow$  & 0.48(0.03) &   0.30(0.01) &  0.27(0.00)   &   0.32(0.01) &      \textbf{0.22(0.00)}  \\
\cmidrule{2-7}
                                                               & Coverage $\uparrow$ & 0.94(0.01) &   \textbf{1.00(0.00)}& \textbf{1.00(0.00)}&  \textbf{1.00(0.00)} & \textbf{1.00(0.00) }         \\
\cmidrule{2-7}
                                                               & ACS $\uparrow$      & 110.56(1.11) & 126.03(0.39) & 126.75(0.18)   & 125.15(0.58) &    \textbf{128.49(0.37) }         \\ 
\midrule
\multirow{5}{*}{4 $\Rightarrow$ 3}                                           & Time $\downarrow$   & 96.15(0.46) & 53.68(1.01) & 55.33(0.82)& 52.26(0.68) &     \textbf{48.88(1.82) } \\ \cmidrule{2-7}              & Overlap $\downarrow$  & 0.40(0.03) &  0.36(0.00) &  0.34(0.01) &  0.38(0.01) &      \textbf{0.33(0.07)}      \\ \cmidrule{2-7}
                                                               & Coverage $\uparrow$ & 0.92(0.01) &  \textbf{1.00(0.00)}& \textbf{1.00(0.00)}& \textbf{1.00(0.00)}&  \textbf{1.00(0.00)}         \\ \cmidrule{2-7}
                                                               & ACS $\uparrow$     & 73.28(1.55) & 84.05(0.57) &  84.08(0.41)  & 83.60(0.41) &     \textbf{84.78(0.77)  }    \\ 
\bottomrule
\end{tabular}
\caption{Performance of different methods with decreased team size on $25 \times 25$ maps in the grid-based simulator.}
\label{tab: varying}
\vspace{-5mm}
\end{table*}
\begin{table*}[ht!]

\centering
\begin{tabular}{ccccccc} 
\toprule
Methods                        & Utility & Nearest & RRT & APF & MAPPO & {\name}  \\ 
\midrule
Time(s) &    60.25(0.16)    &    38.72(0.12)     &  55.89(0.24)  &   52.64(0.23)  &  28.48(0.12)  &  \textbf{25.61(0.10)}  \\ 
\bottomrule
\end{tabular}
\caption{Running time of different methods when the coverage ratio reaches 100\% in the real-world robot system.}
\vspace{-5mm}
\label{tab: time}
\end{table*}

\section{Experiment Results}

\subsection{Training Details}
In the simulation, every RL policy is trained with $50M$ steps in the grid-based simulator and $100M$ steps in Habitat over 3 random seeds. All results are averaged over a total of 300 testing episodes (100 episodes per random seed). As for real-world testing, we randomly generate $10$ maps of size $15 \times 15$ and test $5$ times for each map.
In synchronous action-making cases, agents perform action-making at the same time and wait for all other agents to finish. In asynchronous action-making cases, agents do not wait for others and perform both macro and atomic actions independently.

\subsection{Evaluation Metrics}
The most important metric in our experiment is \emph{Time}, which is the running time for the agents to reach a $C\%$ coverage ratio. We report wall-clock time in the real world, and report an estimated statistical running time in simulation: turning left or right takes $0.5s$; stepping forward takes $1s$. Policy inference time is fixed to $0.1s$ for both RL and planning-based methods thus the results can better reflect the difference between asynchronous and synchronous settings. 

We also consider 3 additional statistics metrics to capture different characteristics of a particular exploration strategy. 
These metrics are only for analysis, and we primarily focus on \textit{Time} as our performance criterion. 
\begin{itemize}
    \item \emph{Accumulative Coverage Score} (ACS): The overall exploration progress throughout an episode computed as 
    $A_T=\int_{t=0}^T Ratio^t$, where $T$ is the max running time. Higher \emph{ACS} implies faster exploration. 
    
    \item \emph{Coverage}: the \emph{final} ratio of explored area when an episode terminates. Higher implies more exhaustive exploration. 
    
    \item \emph{Overlap}: the ratio of the overlapped region explored by \emph{multiple} agents to the current explored area when C\% coverage is reached. Lower \emph{Overlap} implies better credit assignment.
\end{itemize}

All metrics are calculated with the running time $t$, i.e., the estimated statistical time in simulation and wall-clock time in the real world. 
Each score is reported as "mean (standard deviation)".
\vspace{-1mm}
\subsection{Baselines}
We consider 4 popular planning-based competitors, including a utility-maximizing method (\emph{Utility})~\cite{utility}, a search-based nearest-frontier method (\emph{Nearest})~\cite{yamauchi1997frontier}, a rapid-exploring-random-tree-based method (\emph{RRT})~\cite{RRT}, and an artificial potential field method (\emph{APF})~\cite{APF} which applies resistance forces among agents as a cooperation mechanism. Note that APF is a multi-agent baseline while the other three are commonly used for single-agent tasks. Moreover, all baselines use global information to do planning after every macro action. Different from {\name}, they are not learning-based and are all designed for asynchronous execution. 

\vspace{-1mm}
\subsection{Grid-Based Scenario}

\subsubsection{\textbf{Main Results}}
Experiment results with 2 agents in the grid-based simulator under synchronous and asynchronous training are provided in Table~\ref{tab: sync-async}. In both settings, {\name} outperforms planning-based baselines with $\ge 10\%$ less \textit{Time}, full \textit{Coverage}, and higher \textit{ACS}. Although APF encourages cooperation, its \textit{Overlap} is still higher than {\name}, demonstrating {\name}'s superiority in discovering efficient cooperation strategies. Comparing {\name} with MAPPO, which is trained in a synchronous manner, {\name} demonstrates similar $ACS$ to MAPPO with less \textit{Time} and \textit{Overlap}, which indicates the robustness of {\name} to realistic execution with randomized action delay. Results of 3 agents can be found in appendix D.

\vspace{-1mm}
\subsubsection{\textbf{Generalization to Agent Lost}}
We further consider another setting where the team size decreases within an episode on map size $25 \times 25$ to emulate the real-world scenarios with hardware failure and to examine whether our learned policies can generalize to these extreme cases during execution. ``$N_1\Rightarrow N_2$'' denotes a scenario with $N_1$ agents at the beginning and only $N_2$ agents alive after $50\%$ coverage. As shown in Table~\ref{tab: varying},  
{\name} demonstrates $10\%$ less \textit{Time} than other baselines and obtains the highest \textit{ACS} and lowest \textit{Overlap}, indicating {\name}'s effective zero-shot adaptation to extreme situations where some agents go offline. 
\vspace{-2mm}
\subsubsection{\textbf{Real-World Robot System}}
In this part, we present the running time of different methods with 2 agents in real-world exploration tasks on $15 \times 15$ maps, which are running in an asynchronous manner. The deployment pipeline is described in Sec.~\ref{deploy}. As shown in Table~\ref{tab: time}, two RL-based methods, MAPPO and {\name}, outperform the planning-based baselines with a large margin according to the total exploration time. In particular, {\name} reduces 33.86\% real-world exploration time than the fastest planning-based method \emph{Nearest}. Besides, {\name} reduces 10.07\% running time compared with MAPPO, proving that combining action-delay randomization with Async-MAPPO indeed improves the efficiency of multi-agent exploration.

\vspace{-2mm}

\subsection{Habitat Results}

\subsubsection{\textbf{Main Results}}
We extend {\name} to a vision-based environment, Habitat. Table~\ref{tab:habitat} shows the performance of different methods under 2-agent asynchronous action-making settings. Despite having higher \textit{Overlap} due to more exhaustive exploration, {\name} outperforms planning-based baselines with $\ge 28\%$ less \textit{Time}, higher \textit{Coverage} and \textit{ACS}. Compared with synchronous MAPPO, {\name} still shows higher \textit{Coverage} and \textit{ACS} with less \textit{Time}, demonstrating the effectiveness of {\name} in more complicated vision-based tasks.

\vspace{-1mm}
\subsubsection{\textbf{Generalization to Agent Lost}}
We also consider the setting of decreased team sizes in Habitat, and we follow the same experimental setup as for the grid-based simulations. Table~\ref{tab:habitat-varying} shows the performance of different methods with decreased team size (2 $\Rightarrow$ 1). {\name} demonstrates $5.3\%$ less \textit{Time} than other baselines and obtains the highest \textit{Coverage} and \textit{ACS} with comparable \textit{Overlap}, which indicates the  {\name}'s ability to generalize to agent lost.
\begin{table}[ht!]
\vspace{-2mm}
\centering
\begin{tabular}{cccccc} 
\toprule
Methods    & Time $\downarrow$ & Overlap $\downarrow$  & Coverage $\uparrow$  & ACS  $\uparrow$      \\ 
\midrule
Utility  & 273.83\scriptsize{(37.80)} & 0.84\scriptsize{(0.03)} & 0.83\scriptsize{(0.08)} & 186.17\scriptsize{(17.43)}  \\ 
\midrule
Nearest  & 220.25\scriptsize{(30.23)} & \textbf{0.59\scriptsize{(0.04)}} & 0.94\scriptsize{(0.03)} & 180.05\scriptsize{(7.37)} \\ 
\midrule
RRT  &177.29\scriptsize{(17.16)} & 0.63\scriptsize{(0.06)} & 0.97\scriptsize{(0.02)} & 187.35\scriptsize{(6.86)} \\ 
\midrule
APF & 218.45\scriptsize{(24.64)} & 0.67\scriptsize{(0.04)} & 0.94\scriptsize{(0.02)} & 188.62\scriptsize{(7.67)}   \\ 
\midrule
MAPPO & 133.23\scriptsize{(17.72)} & 0.68\scriptsize{(0.09)} & 0.97\scriptsize{(0.01)} & 201.33\scriptsize{(7.98)}\\
\midrule
{\name} &\textbf{127.62\scriptsize{(8.55)}} & 0.78\scriptsize{(0.07)} & \textbf{0.98\scriptsize{(0.01)}} & \textbf{213.81\scriptsize{(9.33)}}  \\
\bottomrule
\end{tabular}
\caption{Performance of different methods under 2-agent asynchronous action-making settings in Habitat.}
\vspace{-5mm}
\label{tab:habitat}
\end{table}
\begin{table}[htp]
\vspace{-1mm}
\centering
\begin{tabular}{cccccc} 
\toprule
Methods    & Time $\downarrow$ & Overlap $\downarrow$  & Coverage $\uparrow$  & ACS  $\uparrow$      \\ 
\midrule
Utility  & 281.09\scriptsize{(32.25)} & 0.48\scriptsize{(0.05)} & 0.84\scriptsize{(0.08)} & 153.02\scriptsize{(12.38)} \\ 
\midrule
Nearest  & 309.76\scriptsize{(9.83)} & 0.40\scriptsize{(0.03)} & 0.85\scriptsize{(0.05)} & 149.43\scriptsize{(4.55)}  \\ 
\midrule
RRT  & 260.31\scriptsize{(27.92)} & \textbf{0.35\scriptsize{(0.03)}} & \textbf{0.92\scriptsize{(0.02)}} & 155.23\scriptsize{(6.04)} \\ 
\midrule
APF & 309.88\scriptsize{(6.63)} & 0.42\scriptsize{(0.01)} & 0.79\scriptsize{(0.01)} & 143.54\scriptsize{(0.83)}   \\ 
\midrule
MAPPO &262.92\scriptsize{(19.84)} & \textbf{0.35\scriptsize{(0.04)}} & 0.90\scriptsize{(0.03)} & 160.90\scriptsize{(6.68)} \\
\midrule
{\name} &\textbf{246.38\scriptsize{(19.26)}} & 0.36\scriptsize{(0.03)} & \textbf{0.92\scriptsize{(0.03)}} & \textbf{164.32\scriptsize{(8.23)}}   \\
\bottomrule
\end{tabular}
\caption{Performance of different methods with decreased team size in Habitat.}
\vspace{-5mm}
\label{tab:habitat-varying}
\end{table}

\vspace{-4mm}
\subsection{Ablation Studies}
In this section, we analyze the sensitivity of communication size and action-delay randomization based on the grid-like simulator through ablation studies.

\subsubsection{\textbf{Sensitivity Analysis of Communication Size}}
We study the exploration performances in different communication traffic scenarios, including:
\begin{itemize}
    \item \textbf{No Comm.}: The attention-based relation encoder is removed. Therefore, agents can only use their own local information to perform macro actions. This is the lower bound of different communication traffic.
    \item \textbf{Comm. (0.25x)}: The number of channels output by the CNN local feature extractor is set to $1$, which is a quarter of the original $4$ channels. 
    \item \textbf{Comm. (0.5x)}: The number of CNN local feature extractor output channels is set to $2$.
    \item \textbf{Perf. Comm.}: Agents use merged observation from all the agents as the input of the CNN local feature extractor. 
\end{itemize}

Table~\ref{tab:commu} summarizes the performances on different communication traffic with 2 agents on $25 \times 25$ maps. 
More communication between agents generally leads to better exploration efficiency, as is shown by the decreasing \textit{Time} and increasing \textit{ACS} from ``No Comm.'' to ``Comm. (0.25x)'', ``Comm. (0.5x)'' and ``Perf. Comm.''. Moreover, the behavior metric \textit{Overlap} in these four scenarios shows better cooperation efficiency with more communication. Note that {\name} performs even better than ``Perf. Comm.'' with strictly less communication, demonstrating the effectiveness of the feature embedding extracted from our CNN policy for decision-making. 

\begin{table}[htp]
\vspace{-1mm}
\centering
\begin{tabular}{cccccc} 
\toprule
Methods    & Time $\downarrow$ & Overlap $\downarrow$  & Coverage $\uparrow$  & ACS  $\uparrow$      \\ 
\midrule
No Comm.  & 159.26\scriptsize{(2.18)} & 0.37\scriptsize{(0.01)}  & 0.93\scriptsize{(0.01)}&151.87\scriptsize{(1.82)}  \\ 
\midrule
Comm. \scriptsize{(0.25x)}  &110.92\scriptsize{(1.33)} & 0.11\scriptsize{(0.01)}  & 0.99\scriptsize{(0.00)} & 167.60\scriptsize{(0.71)} \\ 
\midrule
Comm. \scriptsize{(0.5x)} & 83.77\scriptsize{(1.38)}& 0.09\scriptsize{(0.00)} & \textbf{1.00\scriptsize{(0.00)}} & 170.90\scriptsize{(0.60)}   \\ 
\midrule
Perf. Comm. &75.62\scriptsize{(0.84)}  &\textbf{0.06\scriptsize{(0.01)}}& \textbf{1.00\scriptsize{(0.00)}} & 173.15\scriptsize{(0.53)} \\
\midrule
{\name} &\textbf{74.36\scriptsize{(2.93)}}  &\textbf{0.06\scriptsize{(0.00)}}  & \textbf{1.00\scriptsize{(0.00)}}&  \textbf{173.16\scriptsize{(0.72)}}   \\
\bottomrule
\end{tabular}
\caption{Performance with different communication traffic.}
\vspace{-5mm}
\label{tab:commu}
\end{table} 

\begin{table}[htp]
\vspace{-5mm}
\centering
\begin{tabular}{ccccc} 
\toprule
Intervals    & Time $\downarrow$ & Overlap $\downarrow$  & Coverage $\uparrow$  & ACS  $\uparrow$      \\ 
\midrule
Rand (1-10)  & 60.24\scriptsize{(0.35)}&0.51\scriptsize{(0.00)}&\textbf{1.00\scriptsize{(0.00)}} & 82.44\scriptsize{(0.31)}  \\ 
\midrule
Rand (5-10) &56.41\scriptsize{(1.51)} & 0.47\scriptsize{(0.01)}& \textbf{1.00\scriptsize{(0.00)}} & 83.46\scriptsize{(0.35)}  \\ 
\midrule
Rand (1-5)  & 55.69\scriptsize{(1.23)} &    0.43\scriptsize{(0.01)} &    \textbf{1.00\scriptsize{(0.00)}} &     83.57\scriptsize{(0.45)}  \\ 
\midrule
{\name} (3-5) & \textbf{48.88\scriptsize{(1.82)}}&\textbf{0.33\scriptsize{(0.07)}} & \textbf{1.00\scriptsize{(0.00)}} & \textbf{84.78\scriptsize{(0.77)}}  \\
\bottomrule
\end{tabular}
\caption{Performance of different action-delay intervals.}
\label{tab:delay}
\vspace{-5mm}
\end{table}

\subsubsection{\textbf{Sensitivity Analysis of Action-Delay Randomization}}

We further study the impact of the different random action-delay intervals. Besides the randomization interval stated in Sec.~\ref{action-delay randomization}, we consider 3 different choices of action-delay intervals during training, ``Rand (1-10)'', ``Rand (5-10)'', and ``Rand (1-5)''. ``Rand ($M_1-M_2$)'' means each macro action execution is delayed for a random number of simulation steps uniformly sampled from $[M_1, M_2]$.
We empirically find that these variants have similar performance in most simple test settings, while {\name} outperforms them in some extreme cases. To better illustrate the effect of different action-delay choices, we present the results in the ``$4\Rightarrow 3$'' setting, an extreme scenario with agent loss. As shown in Table~\ref{tab:delay}, {\name} consumes the least \textit{Time} and achieves the highest \textit{ACS}. The results show that action-delay randomization works best with a proper randomization interval, while a large randomization interval adds high uncertainty during training and hurts the final performance.

\section{Conclusion}
To bridge the gap between synchronous simulator and asynchronous action-making process in real-world multi-agent exploration task, we propose a novel real-world multi-robot exploration solution, \emph{Asynchronous Coordination Explorer ({\name})} to tackle this challenge. In {\name}, Multi-agent PPO (MAPPO) is extended to the asynchronous action-making setting for effective training, and an action-delay-randomization technique is applied for better generalization to the real world. Besides, each agent equipped with a team-size-invariant Multi-tower-CNN-based Policy ({\planner}), extracts and broadcasts the low-dimensional feature embedding to accomplish efficient intra-agent communication. Both simulation and real-world results show that {\name} improves $10\%$ exploration efficiency compared with classical approaches in grid-based environments. We also extend {\name} to a vision-based testbed Habitat, where {\name} outperforms planning-based baselines with $\ge 28\%$ less exploration time. Although we aim at the sim-to-real problem caused by multiple agents executing tasks asynchronously, there are still many issues that have not been fully considered, such as communication errors, localization errors, and sensor errors. we leave these issues as our future work.

\section*{ACKNOWLEDGMENT}

This research was supported by National Natural Science Foundation of China (No.U19B2019, 62203257, M-0248), Tsinghua University Initiative Scientific Research Program, Tsinghua-Meituan Joint Institute for Digital Life, Beijing National Research Center for Information Science, Technology (BNRist), and Beijing Innovation Center for Future Chips and 2030 Innovation Megaprojects of China (Programme on New Generation Artificial Intelligence) Grant No. 2021AAA0150000.

\clearpage
\appendix

We would suggest to visit \url{https://sites.google.com/view/ace-aamas} for more information.

\section{Habitat Details}



\subsection{Pipeline}


In Habitat experiments, we use Neural SLAM to represent the scene with a top-down mapping, and thus the explored regions and discovered obstacles are expressed with a top-down 2D mapping. Then the planner, {\planner}, schedules a global goal according to the explored information. Finally, the agent uses a local policy that guides the agent to the chosen global goal.

\subsection{{\planner} Details}

\subsubsection{Input Representation}

In {\planner}, each CNN-based feature extractor's input map, i.e. one feature extractor per agent, is a $240\times 240$ map with 7 channels, including

\begin{itemize}
    \item Obstacle channel: Each pixel value denotes the probability of being an obstacle.
    \item Explored region channel: A probability map for each pixel being explored.
    \item One-hot location channel: The only non-zero grid denotes the position of the agent.
    \item Trajectory channel: This is used to represent the agent's history trace. To reflect time-passing, this channel is updated in an exponentially decaying weight manner. More precisely, an agent's trajectory channel $V^t$ at timestep $t$ is updated as following,
    \begin{equation*}
        \begin{array}{l}V_{x,y}^t=\left\{\begin{array}{l}
        \begin{array}{l}1\quad \quad\quad\text{if agent is near }(x,y)\\\varepsilon V^{t-1}_{x,y}\quad \text{otherwise}\end{array}\end{array} \right.
        \\
        \end{array}
    \end{equation*}
    where the agent is regarded as near $(x,y)$ when the grid-level distance between them is less than $3$.
    \item Three local observations channels: the RGB images of agent-view local observations.
\end{itemize}

All mapping-related channels are transformed into a world-view to save {\planner} from learning to align all agents' information, which might involve rotation and translation. 

\subsubsection{Action Space}

Through {\planner}, every agent chooses a long-term goal (a point) from the whole space. A natural choice is to model the agent's policy as a multi-variable Gaussian distribution to select points from a plane. However, in our exploration setting, an agent's policy could be extremely multi-modal especially during early stage of exploration since many points could induce similar effects on the agent's path. To fix this issue, we adopt a hierarchical design. We first divide the whole map into $8\times 8$ regions, from which the agent chooses a desired region. Then, similar to previous choice, a point in this region is selected as the long-term goal. Formally, the policy of agent $k$, could be described as,
\begin{align*}
    g_r,g_c&\sim ~\text{Cat}(r_{\theta,r}),\text{Car}(r_{\theta,c})\\
    x_l,y_l&\sim ~\mathcal{N}(\mu_{\theta},\Sigma_{\theta})\\
    x'_l=\text{sigmoid}(x_l)&,~~ y'_l=\text{sigmoid}(y_l)\\
    x_g=(g_r+x'_l)/8&,~~ y_g=(g_c+y'_l)/8
\end{align*}
where $\theta$ is the model parameter, $\text{Cat}(r_{\theta,r}),\text{Car}(r_{\theta,c})$ represent two categorical distributions for choosing the region, $g_r, g_c$ are the row and column indexes of the sampled region, $\mu_\theta,\Sigma_\theta$ are the mean and covariance matrix of the Gaussian distribution to choose the local point within the region and $(x_g,y_g)$ is the final sampled long-term goal.

\subsubsection{Network Architecture} 

Our models are trained and implemented using Pytorch~\cite{paszke2017automatic}. We reuse the neural SLAM module and local policy from \cite{singleagent-RL2}, and we briefly summarize their architectures here. Neural SLAM module has two components, a Mapper and a Pose Estimator. The Mapper is composed of ResNet18 convolutional layers, 2 fully-connected layers, and 3 deconvolutional layers. The Pose Estimator consists of 3 convolutional layers and 3 fully connected layers. Similarly, the local policy has Resnet18 convolutional layers, fully-connected layers, and a recurrent GRU layer. 

\begin{table}[ht!]
\centering
\vspace{-1mm}
\caption{CNN Block Hyperparameter in Habitat.}
\label{tab:cnn}
\begin{tabular}{ccccc}
\toprule
Layer &                      Out Channels &         Kernel Size & Stride & Padding \\
\midrule
1 & 32 & 3 & 1 & 1 \\
    2 &  64 & 3 & 1 & 1 \\
3 & 128 & 3 & 1 & 1 \\
    4 & 64 & 3 & 1 & 1 \\
  5 & 32 & 3 & 2 & 1 \\
\bottomrule
\end{tabular}
\vspace{-2mm}
\end{table}

The Multi-tower-CNN-based Policy ({\planner}) has three main components, including CNN-based feature extractors, a transformer-based relation encoder, and an action decoder.
\begin{enumerate}
    \item Each CNN-based feature extractor contains 5 consecutive CNN blocks. Their corresponding parameters are shown in tab.~\ref{tab:cnn}. We use ReLU as the activation function. After each of the front four CNN blocks, we attach a 2D max pooling layer with 2 kernel sizes.
    \item The transformer-based relation encoder is used to better capture spatial information. The attention layer has 4 heads, with 32 dimension sizes for each head.
    \item The action decoder simply uses a CNN projector and linear transformations to turn the feature map output from the transformer-based relation encoder to corresponding logits for Categorical distribution (region head) and means and standard deviations of the Gaussian distribution (point heads). 
\end{enumerate}

The critic also utilizes a similar architecture as {\planner}, except for replacing the action decoder with fully-connected layers to output value predictions. 

\section{Training Details}

\subsection{Reward Function}

We use 3 kinds of team-based rewards, including a coverage reward, a success reward, and an overlap penalty. In the following part, $Ratio^t$ denotes the total coverage ratio at timestep $t$. Let $Exp^t$ be the merged explored map at timestep $t$ and $Exp_k^t$ be the explored map of agent $k$. Ideally, both $Exp^t$ and $Exp^t_k$ can be considered as sets of explored points. Then define $\Delta Exp^t=Exp^t\backslash Exp^{t-1}$ as the newly discovered region at timestep $t$ by the whole team with regard of the merged explored area. Specially, we model an individual's effort by $\Delta Exp^t_k=Exp^t_k\backslash Exp^{t-1}$, that is agent $k$'s contribution at timestep $k$ based on the whole team's previous exploration. Note that $\Delta Exp^t_k$ is not defined based on the agent's previous exploration, i.e. $\Delta Exp^t_k\neq Exp^t_k\backslash Exp^{t-1}_k$. 

\begin{itemize}
    \item \textbf{Coverage Reward:} The coverage reward consists of two parts, a team coverage reward, and an individual coverage reward. The team coverage reward is proportional to the area of the exploration increment $\Delta Exp^t$. The individual coverage reward, as the name suggests, is proportional to the individual contribution, i.e., the area of $\Delta Exp^t_k$. 
    \item \textbf{Success Reward:} Agent $a$ gets a success reward of $Ratio^t$ when $C\%$ coverage rate is reached, which $C=98$ in the grid-based simulator and $C=90$ in Habitat\footnote{Maps in Habitat are harder than in the grid-based simulator, leading to differences in the success rate threshold.}.
    \item \textbf{Overlap Penalty:} The overlap penalty $r_{overlap}$ is designed to encourage agents to reduce repetitive exploration and learn to cooperate with others. 
    \begin{equation*}
        \begin{array}{l}r_{overlap}=\left\{\begin{array}{l}
        \begin{array}{l}-A_{overlap}\times0.01,\;Ratio^t\;<0.9\\0,\;Ratio^t\ge0.9\end{array}\end{array} \right.
        \\
        \end{array},
    \end{equation*}
    where $A_{overlap}$ is the increment of the overlapped explored area between agent $a$ and other agents. The overlapped area between agent $a$ and agent $w$ is $Overlap^t_{a,w}=Exp^t_a\cap Exp^t_w$, and
    $A_{overlap} = \sum_ {w\in\{1,\cdots,n\}\backslash\{a\}} Overlap^t_{a,w}\backslash Overlap^{t-1}_{a,w}$.
    
\end{itemize}
The final team-based reward is simply the sum of all these terms. In Habitat, all the explored and obstacle maps are represented under discretization of $5cm$, and all the area computations are taken in $m^2$.

\subsection{Hyperparameters}

The hyperparameters for Async-MAPPO are as shown in Table~\ref{tab:hyperparameter}.

\begin{table}[bt]
\centering
\begin{tabular}{cc}
\toprule
common hyperparameters      & value  \\
\midrule
gradient clip norm          & 10.0 \\
GAE lambda                   & 0.95    \\              
gamma                      & 0.99 \\
value loss                        & huber loss \\
huber delta                 & 10.0   \\
mini batch size           & batch size {/} mini-batch  \\
optimizer                & Adam       \\
optimizer epsilon            & 1e-5       \\
weight decay             & 0          \\
network initialization  & Orthogonal \\
use reward normalization   &True \\
use feature normalization   &True \\
learning rate & 2.5e-5 \\
\bottomrule
\end{tabular}
\caption{Async-MAPPO hyperparameters}
\label{tab:hyperparameter}
\end{table}

\section{Planning-based Baselines}

We demonstrate some details about the 5 planning-based baselines here.

\textbf{Utility:} A method that always chooses frontier that maximizes information gain~\cite{frontier3}.

\textbf{Nearest:} A method that always chooses the nearest frontier as global goal~\cite{frontier2}. The distance to a frontier is computed using the breadth-first search on the occupancy map.

\textbf{APF:} Artificial Potential Field (APF)~\cite{APF} plans a path for each agent based on a computed potential field. The end of the path, which is a frontier, is the selected goal. For every agent, an artificial potential field $F$ is computed in the discretized map, with consideration of distance to frontiers, presence of obstacles, and potential exploration reward. APF also introduces resistance force as a simple mechanism. Finally, the path is generated along the fastest decreasing direction of $F$, starting from the agent's current position. 

\textbf{RRT:} This baseline is adopted from \cite{RRT}. Rapid-exploring Random Tree (RRT) is originally a path-planning algorithm based on random sampling and is used as a frontier detector in \cite{RRT}. After collecting enough frontiers through random exploration, RRT chooses frontier $p$ with the largest utility $u(p)=IG(p)-N(p)$, where $IG(p)$ and $N(p)$ are respectively the normalized information gain and navigation cost of $p$. 

\textbf{Voronoi}~\cite{Voronoi} The voronoi-based method first partitions the map via voronoi partition and assigns components to agents so that each agent owns parts that are closest to it. Then each agent finds its own global goal by finding a frontier point with largest potential as in Utility within its own partition.

In Habitat experiment, to avoid visually blind areas and ensure that selected frontiers are far enough, the area within $2.5m$ from each agent is considered explored when making global planning. The information gain of a frontier $p$ is computed as the number of unexplored grids within $1.5m$ to $p$. All these baselines do re-planning every $15$ environment steps.

Pseudocode of APF is shown in \ref{algo:APF}. Line 6-12 computes the resistance force between every pair of agents where $D$ is the influence radius. In lines 13-18, distance maps starting from cluster centers are computed, and the corresponding reciprocals are added into the potential field so as one agent approaches the frontier, the potential drops. Here $w_c$ is the weight of cluster $c$, which is the number of targets in this cluster. Consequently, an agent would prefer to seek frontiers that are closer and with more neighboring frontiers. Lines 20-25 show the process of finding the fastest potential descending path, at each iteration, the agent moves to the cell with the smallest potential among all neighboring ones. $T$ is the maximum number of iterations, and $C_{repeat}$ is the repeat penalty to avoid agents wandering around cells with the same potentials.

Pseudocode of RRT is shown in Algo. \ref{algo:RRT}. In each iteration, a random point $p$ is drawn and a new node $t$ is generated by expanding from $s$ to $p$ with distance $L$, where $s$ is the closest tree node to $p$. If segment $(s,t)$ has no collision with obstacles in $M$, $t$ is inserted into the target list or the tree according to whether $t$ is in the unexplored area or not. Finally, the goal is chosen from the target list with the largest utility $u(c) =IG(c)-N(c)$ where $IG(c)$ is the information gain and $N(c)$ is the navigation cost. $IG(c)$ is computed by the number of unexplored grids within $1.5m$ to $c$, as mentioned above. $N(c)$ is computed as the Euclidean distance between the agent location and point $c$. To keep these two values at the same scale, we normalize $IG(\cdot)$ and $N(\cdot)$ to $[0,1]$ w.r.t all cluster centers.

\begin{algorithm}
    \caption{Rapid-exploring Random Tree}
    \label{algo:RRT}
    \begin{algorithmic}[1]
        \REQUIRE Map $M$ and agent location $loc$.
        \ENSURE Selected frontier goal
        \STATE $NodeList\leftarrow \{loc\}, Targets\leftarrow\{\}$
        \STATE $i\leftarrow 0$
        \WHILE{$i<T$ and $|Targets|<N_{target}$}
            \STATE $i\leftarrow i+1$
            \STATE $p\leftarrow $ a random point
            \STATE $s\leftarrow \arg\min_{u\in NodeList}||u-p||_2$
            \STATE $t\leftarrow Steer(s, p, L)$
            \IF{$No\_Collision(M, s, t)$}
                \IF{$t$ lies in unexplored area}
                    \STATE $Targets\leftarrow Targets + \{t\}$
                \ELSE
                    \STATE $NodeList\leftarrow NodeList + \{t\}$
                \ENDIF
            \ENDIF
        \ENDWHILE
        \STATE $C\leftarrow $ clusters of points in $Targets$.
        \STATE $goal\leftarrow \arg\min_{c\in C} IG(c) - N(c)$
        \RETURN $goal$
    \end{algorithmic}
\end{algorithm}

\begin{algorithm}
    \caption{Artificial Potential Field(APF)}
    \label{algo:APF}
    \begin{algorithmic}[1]
        \REQUIRE Map $M$, number of agents $n$ and agent locations $loc_{1}\dots loc_{n}$.
        \ENSURE Selected goals
        \STATE $P\leftarrow$ frontiers in $M$
        \STATE $C\leftarrow$ clusters of frontiers $P$
        \STATE $goals\leftarrow$ an empty list
        \FOR{$i=1\rightarrow n$}
            \STATE $F\leftarrow$ zero potential field, i.e., a 2d array
            \FOR{$j=1\rightarrow n$}
                \FOR{empty grid $p\in M$}
                    \IF{$j\neq i$ and $||p-loc_j||_2<D$}
                        \STATE $F_p \leftarrow F_p + k_D\cdot(D-||p-loc_j||_2)$
                    \ENDIF
                \ENDFOR
            \ENDFOR
            \FOR{$c\in C$}
                \STATE Run breadth-first search to compute distance map $dis$ starting from $c$
                \FOR{empty grid $p\in M$}
                    \STATE $F_p\leftarrow F_p - dis_p^{-1}\cdot w_c$
                \ENDFOR
            \ENDFOR
            \STATE $u\leftarrow loc_i, cnt\leftarrow 0$
            \WHILE{$u\notin M$ and $F_u$ is not a local minima and $cnt<T$}
                \STATE $cnt\leftarrow cnt+1$
                \STATE $F_u\leftarrow F_u+C_{repeat}$
                \STATE $u\leftarrow\arg\min_{v\in Neigh(u)}F_v$
            \ENDWHILE
            \STATE{append $u$ to the end of $goals$}
        \ENDFOR
        \RETURN $goals$
    \end{algorithmic}
\end{algorithm}

\section{Additional Results}


\subsection{Ablation Studies in Habitat}

We conduct ablation studies on {\planner} and report the training ACS performances on two maps in Habitat.

\subsubsection{{\planner} w.o. RE} We consider the {\planner} without the relation encoder. The output feature maps from the CNN-based feature extractors are channel-wise concatenated and directly fed into the action decoder.
\subsubsection{{\planner} w.o. AD} We remove discrete heads from the action decoder so that the global goal is directly generated via two Gaussian action distributions.

\begin{figure}[ht!]
\vspace{-3mm}
\captionsetup{justification=centering}
	\centering
    \subfigure
	{
        {
        \includegraphics[width=0.24\textwidth]{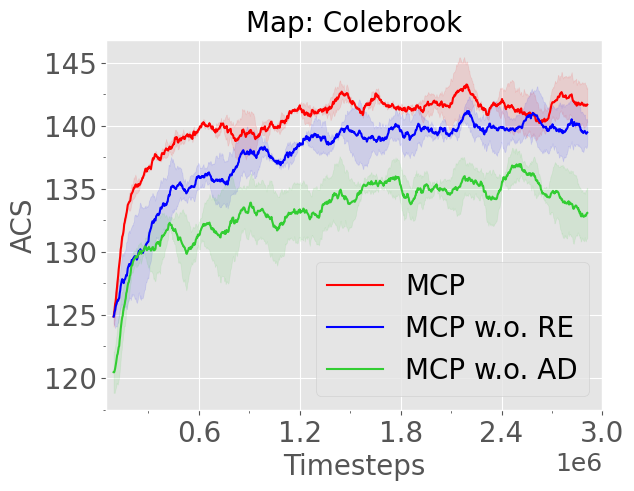}
        \includegraphics[width=0.24\textwidth]{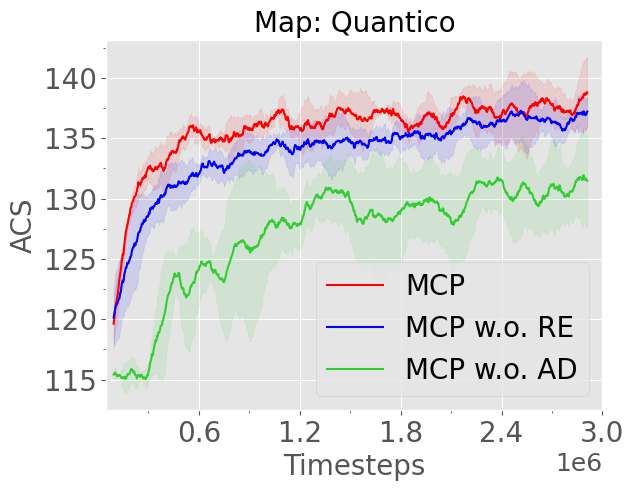}
        }
    }
    \vspace{-5mm}
	\centering \caption{Ablation studies on {\planner} components.}
\label{fig:ab-mcp}
\vspace{-3mm}
\end{figure}
\begin{table*}[ht]
\vspace{-5mm}
\centering
\begin{tabular}{cccccc|cccc} 
\toprule
\multirow{2}{*}{Map Size} & \multirow{2}{*}{Methods} & \multicolumn{4}{c|}{Synchronous Action Making}         & \multicolumn{4}{c}{Asynchronous Action Making}                                        \\ 
\cmidrule{3-10}
                         &                         & Time $\downarrow$  & Overlap $\downarrow$ & Coverage $\uparrow$ & ACS $\uparrow$ & Time   $\downarrow$ & Overlap $\downarrow$ & Coverage $\uparrow$ & ACS $\uparrow$  \\ 
\midrule
\multirow{6}{*}{$25 \times25$}                     & Utility                       & 144.13(1.23)       &0.44(0.05)           & 0.94(0.01)          & 108.0(1.3)      &137.2(3.3)        & 0.5(0.0)          & 0.95(0.01)          & 112.53(0.88)    \\
\cmidrule{2-10}
                                           & Nearest                       & 79.55(0.94)        & 0.33(0.02)         & \textbf{1.00(0.00)}          & 125.54(0.39)        &64.65(2.76)        & 0.32(0.02)          & \textbf{1.00(0.00)}          & 129.6(0.6)     \\

\cmidrule{2-10}
                                           & RRT                          & 80.42(1.86)         &0.36(0.02)           & \textbf{1.00(0.00)}          & 126.89(0.50)      & 70.64(2.02)       & 0.33(0.05)       & \textbf{1.00(0.00)}        & 130.20(0.53)      \\ 
\cmidrule{2-10}
                                           & APF                           & 75.67(1.94)         & 0.31(0.03)          & \textbf{1.00(0.00)}         & 124.98(0.33)      & 63.26(2.82)        & 0.33(0.06)           & \textbf{1.00(0.00)}          & 129.02(0.41)     \\ 
\cmidrule{2-10}
                                           & Voronoi                     & 80.63(1.89)      & 0.23(0.04)         & \textbf{1.00(0.00)}          & 126.23(0.32)        & 68.58(1.53)      &0.23(0.01)          & \textbf{1.00(0.00)}        & 129.57(0.13)    \\   
\cmidrule{2-10}
                                           & MAPPO                     &  64.8(0.33)        &   0.13(0.02)       &\textbf{1.00(0.00)}           &129.7(0.14)    &       57.1(0.02)            &  0.12(0.01)           & \textbf{1.00(0.00)}    &     132.03(0.10)          \\

\cmidrule{2-10}
                                           & {\name}             &  \textbf{60.7(0.02)} &  \textbf{0.12(0.01)}&   \textbf{ 1.00(0.00)}&   \textbf{131.23(0.02)} &  \textbf{55.4(0.03)} &  \textbf{0.10(0.02) } &\textbf{1.00(0.00)}
                   &\textbf{ 134.7(0.02)  }         \\
             
\bottomrule
\end{tabular}
\caption{Performances of different methods under 3-agent synchronous and asynchronous settings in the grid-based simulator.}
\label{tab: sync-async-3}
\vspace{-5mm}
\end{table*}

As shown in Fig.~\ref{fig:ab-mcp}, the full {\planner} module produces both the highest ACS while \emph{MCP w.o. AD} produces the lowest ACS. This suggests that a simple Gaussian representation of actions may not be able to fully capture the distribution of good global goals, which can be highly multi-modal in the early exploration stage. \emph{MCP w.o. RE} performs slightly worse than the full {\planner}, indicating that the relation encoder could encourage cooperation and improve exploration efficiency.

\subsection{3-agent Results}

We additionally report the result of 3 agents in a map with size $25\times 25$ in the grid-based simulator under both synchronous and asynchronous settings, shown in Tab.~\ref{tab: sync-async-3}. Among all methods, ACE achieves the best exploration efficiency, with the lowest time, overlap ratio and the highest ACS.



\clearpage
\bibliographystyle{ACM-Reference-Format} 
\bibliography{reference}


\begin{thebibliography}{69}


\ifx \showCODEN    \undefined \def \showCODEN     #1{\unskip}     \fi
\ifx \showDOI      \undefined \def \showDOI       #1{#1}\fi
\ifx \showISBNx    \undefined \def \showISBNx     #1{\unskip}     \fi
\ifx \showISBNxiii \undefined \def \showISBNxiii  #1{\unskip}     \fi
\ifx \showISSN     \undefined \def \showISSN      #1{\unskip}     \fi
\ifx \showLCCN     \undefined \def \showLCCN      #1{\unskip}     \fi
\ifx \shownote     \undefined \def \shownote      #1{#1}          \fi
\ifx \showarticletitle \undefined \def \showarticletitle #1{#1}   \fi
\ifx \showURL      \undefined \def \showURL       {\relax}        \fi
\providecommand\bibfield[2]{#2}
\providecommand\bibinfo[2]{#2}
\providecommand\natexlab[1]{#1}
\providecommand\showeprint[2][]{arXiv:#2}

\bibitem[\protect\citeauthoryear{Baker}{Baker}{2020}]%
        {baker2020emergent}
\bibfield{author}{\bibinfo{person}{Bowen Baker}.}
  \bibinfo{year}{2020}\natexlab{}.
\newblock \showarticletitle{Emergent Reciprocity and Team Formation from
  Randomized Uncertain Social Preferences}.
\newblock \bibinfo{journal}{\emph{arXiv preprint arXiv:2011.05373}}
  (\bibinfo{year}{2020}).
\newblock


\bibitem[\protect\citeauthoryear{Bresson, Alsayed, Yu, and Glaser}{Bresson
  et~al\mbox{.}}{2017}]%
        {autonomousdriving}
\bibfield{author}{\bibinfo{person}{Guillaume Bresson}, \bibinfo{person}{Zayed
  Alsayed}, \bibinfo{person}{Li Yu}, {and} \bibinfo{person}{S{\'e}bastien
  Glaser}.} \bibinfo{year}{2017}\natexlab{}.
\newblock \showarticletitle{Simultaneous localization and mapping: A survey of
  current trends in autonomous driving}.
\newblock \bibinfo{journal}{\emph{IEEE Transactions on Intelligent Vehicles}}
  \bibinfo{volume}{2}, \bibinfo{number}{3} (\bibinfo{year}{2017}),
  \bibinfo{pages}{194--220}.
\newblock


\bibitem[\protect\citeauthoryear{Burgard, Moors, Stachniss, and
  Schneider}{Burgard et~al\mbox{.}}{2005}]%
        {frontier3}
\bibfield{author}{\bibinfo{person}{Wolfram Burgard}, \bibinfo{person}{Mark
  Moors}, \bibinfo{person}{Cyrill Stachniss}, {and} \bibinfo{person}{Frank~E
  Schneider}.} \bibinfo{year}{2005}\natexlab{}.
\newblock \showarticletitle{Coordinated multi-robot exploration}.
\newblock \bibinfo{journal}{\emph{IEEE Transactions on robotics}}
  \bibinfo{volume}{21}, \bibinfo{number}{3} (\bibinfo{year}{2005}),
  \bibinfo{pages}{376--386}.
\newblock


\bibitem[\protect\citeauthoryear{Chaplot, Gandhi, Gupta, Gupta, and
  Salakhutdinov}{Chaplot et~al\mbox{.}}{2020}]%
        {ans}
\bibfield{author}{\bibinfo{person}{Devendra~Singh Chaplot},
  \bibinfo{person}{Dhiraj Gandhi}, \bibinfo{person}{Saurabh Gupta},
  \bibinfo{person}{Abhinav Gupta}, {and} \bibinfo{person}{Ruslan
  Salakhutdinov}.} \bibinfo{year}{2020}\natexlab{}.
\newblock \showarticletitle{Learning to explore using active neural slam}. In
  \bibinfo{booktitle}{\emph{International Conference on Learning
  Representations}}. ICLR.
\newblock


\bibitem[\protect\citeauthoryear{Chen, Xu, Li, and Zhao}{Chen
  et~al\mbox{.}}{2021a}]%
        {CHEN2021119}
\bibfield{author}{\bibinfo{person}{Baiming Chen}, \bibinfo{person}{Mengdi Xu},
  \bibinfo{person}{Liang Li}, {and} \bibinfo{person}{Ding Zhao}.}
  \bibinfo{year}{2021}\natexlab{a}.
\newblock \showarticletitle{Delay-aware model-based reinforcement learning for
  continuous control}.
\newblock \bibinfo{journal}{\emph{Neurocomputing}}  \bibinfo{volume}{450}
  (\bibinfo{year}{2021}), \bibinfo{pages}{119--128}.
\newblock
\showISSN{0925-2312}
\urldef\tempurl%
\url{https://doi.org/10.1016/j.neucom.2021.04.015}
\showDOI{\tempurl}


\bibitem[\protect\citeauthoryear{Chen, Zhang, Xu, Ma, Yang, Song, Wang, and
  Wu}{Chen et~al\mbox{.}}{2021b}]%
        {chen2021variational}
\bibfield{author}{\bibinfo{person}{Jiayu Chen}, \bibinfo{person}{Yuanxin
  Zhang}, \bibinfo{person}{Yuanfan Xu}, \bibinfo{person}{Huimin Ma},
  \bibinfo{person}{Huazhong Yang}, \bibinfo{person}{Jiaming Song},
  \bibinfo{person}{Yu Wang}, {and} \bibinfo{person}{Yi Wu}.}
  \bibinfo{year}{2021}\natexlab{b}.
\newblock \bibinfo{title}{Variational Automatic Curriculum Learning for
  Sparse-Reward Cooperative Multi-Agent Problems}.
\newblock
\newblock
\showeprint[arxiv]{2111.04613}~[cs.LG]


\bibitem[\protect\citeauthoryear{Chen, Gupta, and Gupta}{Chen
  et~al\mbox{.}}{2019}]%
        {singleagent-RL1}
\bibfield{author}{\bibinfo{person}{Tao Chen}, \bibinfo{person}{Saurabh Gupta},
  {and} \bibinfo{person}{Abhinav Gupta}.} \bibinfo{year}{2019}\natexlab{}.
\newblock \showarticletitle{Learning exploration policies for navigation}. In
  \bibinfo{booktitle}{\emph{International Conference on Learning
  Representations}}. ICLR.
\newblock


\bibitem[\protect\citeauthoryear{Chevalier-Boisvert, Willems, and
  Pal}{Chevalier-Boisvert et~al\mbox{.}}{2018}]%
        {gym_minigrid}
\bibfield{author}{\bibinfo{person}{Maxime Chevalier-Boisvert},
  \bibinfo{person}{Lucas Willems}, {and} \bibinfo{person}{Suman Pal}.}
  \bibinfo{year}{2018}\natexlab{}.
\newblock \bibinfo{title}{Minimalistic Gridworld Environment for OpenAI Gym}.
\newblock
  \bibinfo{howpublished}{\url{https://github.com/maximecb/gym-minigrid}}.
\newblock


\bibitem[\protect\citeauthoryear{Chu and Ye}{Chu and Ye}{2017}]%
        {DBLP:journals/corr/abs-1710-00336}
\bibfield{author}{\bibinfo{person}{Xiangxiang Chu} {and}
  \bibinfo{person}{Hangjun Ye}.} \bibinfo{year}{2017}\natexlab{}.
\newblock \showarticletitle{Parameter Sharing Deep Deterministic Policy
  Gradient for Cooperative Multi-agent Reinforcement Learning}.
\newblock \bibinfo{journal}{\emph{CoRR}}  \bibinfo{volume}{abs/1710.00336}
  (\bibinfo{year}{2017}).
\newblock
\showeprint[arXiv]{1710.00336}


\bibitem[\protect\citeauthoryear{Cohen}{Cohen}{1996}]%
        {multi-classical2}
\bibfield{author}{\bibinfo{person}{William~W Cohen}.}
  \bibinfo{year}{1996}\natexlab{}.
\newblock \showarticletitle{Adaptive mapping and navigation by teams of simple
  robots}.
\newblock \bibinfo{journal}{\emph{Robotics and autonomous systems}}
  \bibinfo{volume}{18}, \bibinfo{number}{4} (\bibinfo{year}{1996}),
  \bibinfo{pages}{411--434}.
\newblock


\bibitem[\protect\citeauthoryear{Czarnecki, Jayakumar, Jaderberg, Hasenclever,
  Teh, Heess, Osindero, and Pascanu}{Czarnecki et~al\mbox{.}}{2018}]%
        {czarnecki2018mix}
\bibfield{author}{\bibinfo{person}{Wojciech Czarnecki},
  \bibinfo{person}{Siddhant Jayakumar}, \bibinfo{person}{Max Jaderberg},
  \bibinfo{person}{Leonard Hasenclever}, \bibinfo{person}{Yee~Whye Teh},
  \bibinfo{person}{Nicolas Heess}, \bibinfo{person}{Simon Osindero}, {and}
  \bibinfo{person}{Razvan Pascanu}.} \bibinfo{year}{2018}\natexlab{}.
\newblock \showarticletitle{Mix \& match agent curricula for reinforcement
  learning}. In \bibinfo{booktitle}{\emph{International Conference on Machine
  Learning}}. PMLR, \bibinfo{pages}{1087--1095}.
\newblock


\bibitem[\protect\citeauthoryear{Ding, Li, Qian, and Chen}{Ding
  et~al\mbox{.}}{2018}]%
        {ding2018hierarchical}
\bibfield{author}{\bibinfo{person}{Wenhao Ding}, \bibinfo{person}{Shuaijun Li},
  \bibinfo{person}{Huihuan Qian}, {and} \bibinfo{person}{Yongquan Chen}.}
  \bibinfo{year}{2018}\natexlab{}.
\newblock \showarticletitle{Hierarchical reinforcement learning framework
  towards multi-agent navigation}. In \bibinfo{booktitle}{\emph{2018 IEEE
  International Conference on Robotics and Biomimetics (ROBIO)}}. IEEE,
  \bibinfo{pages}{237--242}.
\newblock


\bibitem[\protect\citeauthoryear{Dornhege and Kleiner}{Dornhege and
  Kleiner}{2013}]%
        {sample3}
\bibfield{author}{\bibinfo{person}{Christian Dornhege} {and}
  \bibinfo{person}{Alexander Kleiner}.} \bibinfo{year}{2013}\natexlab{}.
\newblock \showarticletitle{A frontier-void-based approach for autonomous
  exploration in 3d}.
\newblock \bibinfo{journal}{\emph{Advanced Robotics}} \bibinfo{volume}{27},
  \bibinfo{number}{6} (\bibinfo{year}{2013}), \bibinfo{pages}{459--468}.
\newblock


\bibitem[\protect\citeauthoryear{Dosovitskiy, Beyer, Kolesnikov, Weissenborn,
  Zhai, Unterthiner, Dehghani, Minderer, Heigold, Gelly,
  et~al\mbox{.}}{Dosovitskiy et~al\mbox{.}}{2020}]%
        {attention1}
\bibfield{author}{\bibinfo{person}{Alexey Dosovitskiy}, \bibinfo{person}{Lucas
  Beyer}, \bibinfo{person}{Alexander Kolesnikov}, \bibinfo{person}{Dirk
  Weissenborn}, \bibinfo{person}{Xiaohua Zhai}, \bibinfo{person}{Thomas
  Unterthiner}, \bibinfo{person}{Mostafa Dehghani}, \bibinfo{person}{Matthias
  Minderer}, \bibinfo{person}{Georg Heigold}, \bibinfo{person}{Sylvain Gelly},
  {et~al\mbox{.}}} \bibinfo{year}{2020}\natexlab{}.
\newblock \showarticletitle{An image is worth 16x16 words: Transformers for
  image recognition at scale}.
\newblock \bibinfo{journal}{\emph{arXiv preprint arXiv:2010.11929}}
  (\bibinfo{year}{2020}).
\newblock


\bibitem[\protect\citeauthoryear{Duan, Andrychowicz, Stadie, Ho, Schneider,
  Sutskever, Abbeel, and Zaremba}{Duan et~al\mbox{.}}{2017}]%
        {duan2017one}
\bibfield{author}{\bibinfo{person}{Yan Duan}, \bibinfo{person}{Marcin
  Andrychowicz}, \bibinfo{person}{Bradly~C Stadie}, \bibinfo{person}{Jonathan
  Ho}, \bibinfo{person}{Jonas Schneider}, \bibinfo{person}{Ilya Sutskever},
  \bibinfo{person}{Pieter Abbeel}, {and} \bibinfo{person}{Wojciech Zaremba}.}
  \bibinfo{year}{2017}\natexlab{}.
\newblock \showarticletitle{One-Shot Imitation Learning}. In
  \bibinfo{booktitle}{\emph{NIPS}}.
\newblock


\bibitem[\protect\citeauthoryear{Foerster, Assael, de~Freitas, and
  Whiteson}{Foerster et~al\mbox{.}}{2016}]%
        {NIPS2016_c7635bfd}
\bibfield{author}{\bibinfo{person}{Jakob Foerster},
  \bibinfo{person}{Ioannis~Alexandros Assael}, \bibinfo{person}{Nando de
  Freitas}, {and} \bibinfo{person}{Shimon Whiteson}.}
  \bibinfo{year}{2016}\natexlab{}.
\newblock \showarticletitle{Learning to Communicate with Deep Multi-Agent
  Reinforcement Learning}. In \bibinfo{booktitle}{\emph{Advances in Neural
  Information Processing Systems}}, \bibfield{editor}{\bibinfo{person}{D.~Lee},
  \bibinfo{person}{M.~Sugiyama}, \bibinfo{person}{U.~Luxburg},
  \bibinfo{person}{I.~Guyon}, {and} \bibinfo{person}{R.~Garnett}} (Eds.),
  Vol.~\bibinfo{volume}{29}. \bibinfo{publisher}{Curran Associates, Inc.}
\newblock
\urldef\tempurl%
\url{https://proceedings.neurips.cc/paper/2016/file/c7635bfd99248a2cdef8249ef7bfbef4-Paper.pdf}
\showURL{%
\tempurl}


\bibitem[\protect\citeauthoryear{Godoy, Karamouzas, Guy, and Gini}{Godoy
  et~al\mbox{.}}{2015}]%
        {godoy2015adaptive}
\bibfield{author}{\bibinfo{person}{Julio~E Godoy}, \bibinfo{person}{Ioannis
  Karamouzas}, \bibinfo{person}{Stephen~J Guy}, {and} \bibinfo{person}{Maria
  Gini}.} \bibinfo{year}{2015}\natexlab{}.
\newblock \showarticletitle{Adaptive learning for multi-agent navigation}. In
  \bibinfo{booktitle}{\emph{Proceedings of the 2015 International conference on
  autonomous agents and multiagent systems}}. \bibinfo{pages}{1577--1585}.
\newblock


\bibitem[\protect\citeauthoryear{Hsu, Jeong, Pappas, and Chaudhari}{Hsu
  et~al\mbox{.}}{2020}]%
        {DBLP:journals/corr/abs-2011-08055}
\bibfield{author}{\bibinfo{person}{Christopher~D. Hsu}, \bibinfo{person}{Heejin
  Jeong}, \bibinfo{person}{George~J. Pappas}, {and} \bibinfo{person}{Pratik
  Chaudhari}.} \bibinfo{year}{2020}\natexlab{}.
\newblock \showarticletitle{Scalable Reinforcement Learning Policies for
  Multi-Agent Control}.
\newblock \bibinfo{journal}{\emph{CoRR}}  \bibinfo{volume}{abs/2011.08055}
  (\bibinfo{year}{2020}).
\newblock
\showeprint[arXiv]{2011.08055}


\bibitem[\protect\citeauthoryear{Hu, Lerer, Peysakhovich, and Foerster}{Hu
  et~al\mbox{.}}{2020a}]%
        {pmlr-v119-hu20a}
\bibfield{author}{\bibinfo{person}{Hengyuan Hu}, \bibinfo{person}{Adam Lerer},
  \bibinfo{person}{Alex Peysakhovich}, {and} \bibinfo{person}{Jakob Foerster}.}
  \bibinfo{year}{2020}\natexlab{a}.
\newblock \showarticletitle{“{O}ther-Play” for Zero-Shot Coordination}. In
  \bibinfo{booktitle}{\emph{Proceedings of the 37th International Conference on
  Machine Learning}} \emph{(\bibinfo{series}{Proceedings of Machine Learning
  Research}, Vol.~\bibinfo{volume}{119})},
  \bibfield{editor}{\bibinfo{person}{Hal~Daumé III} {and}
  \bibinfo{person}{Aarti Singh}} (Eds.). \bibinfo{publisher}{PMLR},
  \bibinfo{pages}{4399--4410}.
\newblock


\bibitem[\protect\citeauthoryear{Hu, Niu, Carrasco, Lennox, and Arvin}{Hu
  et~al\mbox{.}}{2020b}]%
        {Voronoi}
\bibfield{author}{\bibinfo{person}{Junyan Hu}, \bibinfo{person}{Hanlin Niu},
  \bibinfo{person}{Joaquin Carrasco}, \bibinfo{person}{Barry Lennox}, {and}
  \bibinfo{person}{Farshad Arvin}.} \bibinfo{year}{2020}\natexlab{b}.
\newblock \showarticletitle{Voronoi-based multi-robot autonomous exploration in
  unknown environments via deep reinforcement learning}.
\newblock \bibinfo{journal}{\emph{IEEE Transactions on Vehicular Technology}}
  \bibinfo{volume}{69}, \bibinfo{number}{12} (\bibinfo{year}{2020}),
  \bibinfo{pages}{14413--14423}.
\newblock


\bibitem[\protect\citeauthoryear{Jaderberg, Czarnecki, Dunning, Marris, Lever,
  Castañeda, Beattie, Rabinowitz, Morcos, Ruderman, Sonnerat, Green, Deason,
  Leibo, Silver, Hassabis, Kavukcuoglu, and Graepel}{Jaderberg
  et~al\mbox{.}}{2019}]%
        {doi:10.1126/science.aau6249}
\bibfield{author}{\bibinfo{person}{Max Jaderberg}, \bibinfo{person}{Wojciech~M.
  Czarnecki}, \bibinfo{person}{Iain Dunning}, \bibinfo{person}{Luke Marris},
  \bibinfo{person}{Guy Lever}, \bibinfo{person}{Antonio~Garcia Castañeda},
  \bibinfo{person}{Charles Beattie}, \bibinfo{person}{Neil~C. Rabinowitz},
  \bibinfo{person}{Ari~S. Morcos}, \bibinfo{person}{Avraham Ruderman},
  \bibinfo{person}{Nicolas Sonnerat}, \bibinfo{person}{Tim Green},
  \bibinfo{person}{Louise Deason}, \bibinfo{person}{Joel~Z. Leibo},
  \bibinfo{person}{David Silver}, \bibinfo{person}{Demis Hassabis},
  \bibinfo{person}{Koray Kavukcuoglu}, {and} \bibinfo{person}{Thore Graepel}.}
  \bibinfo{year}{2019}\natexlab{}.
\newblock \showarticletitle{Human-level performance in 3D multiplayer games
  with population-based reinforcement learning}.
\newblock \bibinfo{journal}{\emph{Science}} \bibinfo{volume}{364},
  \bibinfo{number}{6443} (\bibinfo{year}{2019}), \bibinfo{pages}{859--865}.
\newblock
\urldef\tempurl%
\url{https://doi.org/10.1126/science.aau6249}
\showDOI{\tempurl}


\bibitem[\protect\citeauthoryear{Jia, Hu, Chen, Ren, Lv, Fan, and Zhang}{Jia
  et~al\mbox{.}}{2020}]%
        {jia2020fever}
\bibfield{author}{\bibinfo{person}{Hangtian Jia}, \bibinfo{person}{Yujing Hu},
  \bibinfo{person}{Yingfeng Chen}, \bibinfo{person}{Chunxu Ren},
  \bibinfo{person}{Tangjie Lv}, \bibinfo{person}{Changjie Fan}, {and}
  \bibinfo{person}{Chongjie Zhang}.} \bibinfo{year}{2020}\natexlab{}.
\newblock \showarticletitle{Fever basketball: A complex, flexible, and
  asynchronized sports game environment for multi-agent reinforcement
  learning}.
\newblock \bibinfo{journal}{\emph{arXiv preprint arXiv:2012.03204}}
  (\bibinfo{year}{2020}).
\newblock


\bibitem[\protect\citeauthoryear{Jiang, Dun, Huang, and Lu}{Jiang
  et~al\mbox{.}}{2018}]%
        {dgn}
\bibfield{author}{\bibinfo{person}{Jiechuan Jiang}, \bibinfo{person}{Chen Dun},
  \bibinfo{person}{Tiejun Huang}, {and} \bibinfo{person}{Zongqing Lu}.}
  \bibinfo{year}{2018}\natexlab{}.
\newblock \showarticletitle{Graph convolutional reinforcement learning}.
\newblock \bibinfo{journal}{\emph{arXiv preprint arXiv:1810.09202}}
  (\bibinfo{year}{2018}).
\newblock


\bibitem[\protect\citeauthoryear{Jiang and Lu}{Jiang and Lu}{2018}]%
        {NEURIPS2018_6a8018b3}
\bibfield{author}{\bibinfo{person}{Jiechuan Jiang} {and}
  \bibinfo{person}{Zongqing Lu}.} \bibinfo{year}{2018}\natexlab{}.
\newblock \showarticletitle{Learning Attentional Communication for Multi-Agent
  Cooperation}. In \bibinfo{booktitle}{\emph{Advances in Neural Information
  Processing Systems}}, \bibfield{editor}{\bibinfo{person}{S.~Bengio},
  \bibinfo{person}{H.~Wallach}, \bibinfo{person}{H.~Larochelle},
  \bibinfo{person}{K.~Grauman}, \bibinfo{person}{N.~Cesa-Bianchi}, {and}
  \bibinfo{person}{R.~Garnett}} (Eds.), Vol.~\bibinfo{volume}{31}.
  \bibinfo{publisher}{Curran Associates, Inc.}
\newblock
\urldef\tempurl%
\url{https://proceedings.neurips.cc/paper/2018/file/6a8018b3a00b69c008601b8becae392b-Paper.pdf}
\showURL{%
\tempurl}


\bibitem[\protect\citeauthoryear{Juli{\'a}, Gil, and Reinoso}{Juli{\'a}
  et~al\mbox{.}}{2012}]%
        {utility}
\bibfield{author}{\bibinfo{person}{Miguel Juli{\'a}}, \bibinfo{person}{Arturo
  Gil}, {and} \bibinfo{person}{Oscar Reinoso}.}
  \bibinfo{year}{2012}\natexlab{}.
\newblock \showarticletitle{A comparison of path planning strategies for
  autonomous exploration and mapping of unknown environments}.
\newblock \bibinfo{journal}{\emph{Autonomous Robots}} \bibinfo{volume}{33},
  \bibinfo{number}{4} (\bibinfo{year}{2012}), \bibinfo{pages}{427--444}.
\newblock


\bibitem[\protect\citeauthoryear{Kim, Moon, Hostallero, Kang, Lee, Son, and
  Yi}{Kim et~al\mbox{.}}{2019}]%
        {kim2018learning}
\bibfield{author}{\bibinfo{person}{Daewoo Kim}, \bibinfo{person}{Sangwoo Moon},
  \bibinfo{person}{David Hostallero}, \bibinfo{person}{Wan~Ju Kang},
  \bibinfo{person}{Taeyoung Lee}, \bibinfo{person}{Kyunghwan Son}, {and}
  \bibinfo{person}{Yung Yi}.} \bibinfo{year}{2019}\natexlab{}.
\newblock \showarticletitle{Learning to Schedule Communication in Multi-agent
  Reinforcement Learning}. In \bibinfo{booktitle}{\emph{International
  Conference on Learning Representations}}.
\newblock
\urldef\tempurl%
\url{https://openreview.net/forum?id=SJxu5iR9KQ}
\showURL{%
\tempurl}


\bibitem[\protect\citeauthoryear{Kleiner, Prediger, and Nebel}{Kleiner
  et~al\mbox{.}}{2006}]%
        {rescue}
\bibfield{author}{\bibinfo{person}{Alexander Kleiner}, \bibinfo{person}{Johann
  Prediger}, {and} \bibinfo{person}{Bernhard Nebel}.}
  \bibinfo{year}{2006}\natexlab{}.
\newblock \showarticletitle{RFID technology-based exploration and SLAM for
  search and rescue}. In \bibinfo{booktitle}{\emph{2006 IEEE/RSJ International
  Conference on Intelligent Robots and Systems}}. IEEE,
  \bibinfo{pages}{4054--4059}.
\newblock


\bibitem[\protect\citeauthoryear{Long, Zhou, Gupta, Fang, Wu, and Wang}{Long
  et~al\mbox{.}}{2020}]%
        {epciclr2020}
\bibfield{author}{\bibinfo{person}{Qian Long}, \bibinfo{person}{Zihan Zhou},
  \bibinfo{person}{Abhinav Gupta}, \bibinfo{person}{Fei Fang},
  \bibinfo{person}{Yi Wu}, {and} \bibinfo{person}{Xiaolong Wang}.}
  \bibinfo{year}{2020}\natexlab{}.
\newblock \showarticletitle{Evolutionary Population Curriculum for Scaling
  Multi-Agent Reinforcement Learning}. In
  \bibinfo{booktitle}{\emph{International Conference on Learning
  Representations}}.
\newblock


\bibitem[\protect\citeauthoryear{Loquercio, Kaufmann, Ranftl, Müller, Koltun,
  and Scaramuzza}{Loquercio et~al\mbox{.}}{2021}]%
        {doi:10.1126/scirobotics.abg5810}
\bibfield{author}{\bibinfo{person}{Antonio Loquercio}, \bibinfo{person}{Elia
  Kaufmann}, \bibinfo{person}{René Ranftl}, \bibinfo{person}{Matthias
  Müller}, \bibinfo{person}{Vladlen Koltun}, {and} \bibinfo{person}{Davide
  Scaramuzza}.} \bibinfo{year}{2021}\natexlab{}.
\newblock \showarticletitle{Learning high-speed flight in the wild}.
\newblock \bibinfo{journal}{\emph{Science Robotics}} \bibinfo{volume}{6},
  \bibinfo{number}{59} (\bibinfo{year}{2021}), \bibinfo{pages}{eabg5810}.
\newblock
\urldef\tempurl%
\url{https://doi.org/10.1126/scirobotics.abg5810}
\showDOI{\tempurl}
\showeprint{https://www.science.org/doi/pdf/10.1126/scirobotics.abg5810}


\bibitem[\protect\citeauthoryear{Niu, Paleja, and Gombolay}{Niu
  et~al\mbox{.}}{2021}]%
        {niu2021multi}
\bibfield{author}{\bibinfo{person}{Yaru Niu}, \bibinfo{person}{Rohan Paleja},
  {and} \bibinfo{person}{Matthew Gombolay}.} \bibinfo{year}{2021}\natexlab{}.
\newblock \showarticletitle{Multi-agent graph-attention communication and
  teaming}. In \bibinfo{booktitle}{\emph{Proceedings of the 20th International
  Conference on Autonomous Agents and MultiAgent Systems}}.
  \bibinfo{pages}{964--973}.
\newblock


\bibitem[\protect\citeauthoryear{Omidshafiei, Agha{-}mohammadi, Amato, and
  How}{Omidshafiei et~al\mbox{.}}{2015}]%
        {DBLP:journals/corr/OmidshafieiAAH15}
\bibfield{author}{\bibinfo{person}{Shayegan Omidshafiei},
  \bibinfo{person}{Ali{-}akbar Agha{-}mohammadi}, \bibinfo{person}{Christopher
  Amato}, {and} \bibinfo{person}{Jonathan~P. How}.}
  \bibinfo{year}{2015}\natexlab{}.
\newblock \showarticletitle{Decentralized Control of Partially Observable
  Markov Decision Processes using Belief Space Macro-actions}.
\newblock \bibinfo{journal}{\emph{CoRR}}  \bibinfo{volume}{abs/1502.06030}
  (\bibinfo{year}{2015}).
\newblock
\showeprint[arXiv]{1502.06030}
\urldef\tempurl%
\url{http://arxiv.org/abs/1502.06030}
\showURL{%
\tempurl}


\bibitem[\protect\citeauthoryear{OpenAI, Andrychowicz, Baker, Chociej,
  J{\'{o}}zefowicz, McGrew, Pachocki, Pachocki, Petron, Plappert, Powell, Ray,
  Schneider, Sidor, Tobin, Welinder, Weng, and Zaremba}{OpenAI
  et~al\mbox{.}}{2018}]%
        {DBLP:journals/corr/abs-1808-00177}
\bibfield{author}{\bibinfo{person}{OpenAI}, \bibinfo{person}{Marcin
  Andrychowicz}, \bibinfo{person}{Bowen Baker}, \bibinfo{person}{Maciek
  Chociej}, \bibinfo{person}{Rafal J{\'{o}}zefowicz}, \bibinfo{person}{Bob
  McGrew}, \bibinfo{person}{Jakub~W. Pachocki}, \bibinfo{person}{Jakub
  Pachocki}, \bibinfo{person}{Arthur Petron}, \bibinfo{person}{Matthias
  Plappert}, \bibinfo{person}{Glenn Powell}, \bibinfo{person}{Alex Ray},
  \bibinfo{person}{Jonas Schneider}, \bibinfo{person}{Szymon Sidor},
  \bibinfo{person}{Josh Tobin}, \bibinfo{person}{Peter Welinder},
  \bibinfo{person}{Lilian Weng}, {and} \bibinfo{person}{Wojciech Zaremba}.}
  \bibinfo{year}{2018}\natexlab{}.
\newblock \showarticletitle{Learning Dexterous In-Hand Manipulation}.
\newblock \bibinfo{journal}{\emph{CoRR}}  \bibinfo{volume}{abs/1808.00177}
  (\bibinfo{year}{2018}).
\newblock
\showeprint[arXiv]{1808.00177}
\urldef\tempurl%
\url{http://arxiv.org/abs/1808.00177}
\showURL{%
\tempurl}


\bibitem[\protect\citeauthoryear{Osindero, Czarnecki, Vinyals, Dunning, and
  Kavukcuoglu}{Osindero et~al\mbox{.}}{2017}]%
        {osindero2017population}
\bibfield{author}{\bibinfo{person}{Max Jaderberg Valentin Dalibard~Simon
  Osindero}, \bibinfo{person}{Wojciech~M Czarnecki}, \bibinfo{person}{Jeff
  Donahue Ali Razavi~Oriol Vinyals}, \bibinfo{person}{Tim Green~Iain Dunning},
  {and} \bibinfo{person}{Karen Simonyan Chrisantha Fernando~Koray
  Kavukcuoglu}.} \bibinfo{year}{2017}\natexlab{}.
\newblock \showarticletitle{Population Based Training of Neural Networks}.
\newblock \bibinfo{journal}{\emph{arXiv preprint arXiv:1711.09846}}
  (\bibinfo{year}{2017}).
\newblock


\bibitem[\protect\citeauthoryear{Paszke, Gross, Chintala, Chanan, Yang, DeVito,
  Lin, Desmaison, Antiga, and Lerer}{Paszke et~al\mbox{.}}{2017}]%
        {paszke2017automatic}
\bibfield{author}{\bibinfo{person}{Adam Paszke}, \bibinfo{person}{Sam Gross},
  \bibinfo{person}{Soumith Chintala}, \bibinfo{person}{Gregory Chanan},
  \bibinfo{person}{Edward Yang}, \bibinfo{person}{Zachary DeVito},
  \bibinfo{person}{Zeming Lin}, \bibinfo{person}{Alban Desmaison},
  \bibinfo{person}{Luca Antiga}, {and} \bibinfo{person}{Adam Lerer}.}
  \bibinfo{year}{2017}\natexlab{}.
\newblock \showarticletitle{Automatic differentiation in PyTorch}. In
  \bibinfo{booktitle}{\emph{NIPS-W}}.
\newblock


\bibitem[\protect\citeauthoryear{Peng, Andrychowicz, Zaremba, and Abbeel}{Peng
  et~al\mbox{.}}{2017}]%
        {DBLP:journals/corr/abs-1710-06537}
\bibfield{author}{\bibinfo{person}{Xue~Bin Peng}, \bibinfo{person}{Marcin
  Andrychowicz}, \bibinfo{person}{Wojciech Zaremba}, {and}
  \bibinfo{person}{Pieter Abbeel}.} \bibinfo{year}{2017}\natexlab{}.
\newblock \showarticletitle{Sim-to-Real Transfer of Robotic Control with
  Dynamics Randomization}.
\newblock \bibinfo{journal}{\emph{CoRR}}  \bibinfo{volume}{abs/1710.06537}
  (\bibinfo{year}{2017}).
\newblock
\showeprint[arXiv]{1710.06537}
\urldef\tempurl%
\url{http://arxiv.org/abs/1710.06537}
\showURL{%
\tempurl}


\bibitem[\protect\citeauthoryear{Rubio, Valero, and Llopis-Albert}{Rubio
  et~al\mbox{.}}{2019}]%
        {robots}
\bibfield{author}{\bibinfo{person}{Francisco Rubio}, \bibinfo{person}{Francisco
  Valero}, {and} \bibinfo{person}{Carlos Llopis-Albert}.}
  \bibinfo{year}{2019}\natexlab{}.
\newblock \showarticletitle{A review of mobile robots: Concepts, methods,
  theoretical framework, and applications}.
\newblock \bibinfo{journal}{\emph{International Journal of Advanced Robotic
  Systems}} \bibinfo{volume}{16}, \bibinfo{number}{2} (\bibinfo{year}{2019}),
  \bibinfo{pages}{1729881419839596}.
\newblock


\bibitem[\protect\citeauthoryear{Ryu, Shin, and Park}{Ryu
  et~al\mbox{.}}{2020}]%
        {hama}
\bibfield{author}{\bibinfo{person}{Heechang Ryu}, \bibinfo{person}{Hayong
  Shin}, {and} \bibinfo{person}{Jinkyoo Park}.}
  \bibinfo{year}{2020}\natexlab{}.
\newblock \showarticletitle{Multi-agent actor-critic with hierarchical graph
  attention network}. In \bibinfo{booktitle}{\emph{Proceedings of the AAAI
  Conference on Artificial Intelligence}}, Vol.~\bibinfo{volume}{34}.
  \bibinfo{pages}{7236--7243}.
\newblock


\bibitem[\protect\citeauthoryear{Savinov, Raichuk, Marinier, Vincent,
  Pollefeys, Lillicrap, and Gelly}{Savinov et~al\mbox{.}}{2019}]%
        {singleagent-RL2}
\bibfield{author}{\bibinfo{person}{Nikolay Savinov}, \bibinfo{person}{Anton
  Raichuk}, \bibinfo{person}{Raphaël Marinier}, \bibinfo{person}{Damien
  Vincent}, \bibinfo{person}{Marc Pollefeys}, \bibinfo{person}{Timothy
  Lillicrap}, {and} \bibinfo{person}{Sylvain Gelly}.}
  \bibinfo{year}{2019}\natexlab{}.
\newblock \showarticletitle{Episodic curiosity through reachability}. In
  \bibinfo{booktitle}{\emph{International Conference on Learning
  Representations}}. ICLR.
\newblock


\bibitem[\protect\citeauthoryear{Savva, Kadian, Maksymets, Zhao, Wijmans, Jain,
  Straub, Liu, Koltun, Malik, et~al\mbox{.}}{Savva et~al\mbox{.}}{2019}]%
        {habitat}
\bibfield{author}{\bibinfo{person}{Manolis Savva}, \bibinfo{person}{Abhishek
  Kadian}, \bibinfo{person}{Oleksandr Maksymets}, \bibinfo{person}{Yili Zhao},
  \bibinfo{person}{Erik Wijmans}, \bibinfo{person}{Bhavana Jain},
  \bibinfo{person}{Julian Straub}, \bibinfo{person}{Jia Liu},
  \bibinfo{person}{Vladlen Koltun}, \bibinfo{person}{Jitendra Malik},
  {et~al\mbox{.}}} \bibinfo{year}{2019}\natexlab{}.
\newblock \showarticletitle{Habitat: A platform for embodied ai research}. In
  \bibinfo{booktitle}{\emph{Proceedings of the IEEE/CVF International
  Conference on Computer Vision}}. \bibinfo{pages}{9339--9347}.
\newblock


\bibitem[\protect\citeauthoryear{Sukhbaatar, szlam, and Fergus}{Sukhbaatar
  et~al\mbox{.}}{2016}]%
        {NIPS2016_55b1927f}
\bibfield{author}{\bibinfo{person}{Sainbayar Sukhbaatar},
  \bibinfo{person}{arthur szlam}, {and} \bibinfo{person}{Rob Fergus}.}
  \bibinfo{year}{2016}\natexlab{}.
\newblock \showarticletitle{Learning Multiagent Communication with
  Backpropagation}. In \bibinfo{booktitle}{\emph{Advances in Neural Information
  Processing Systems}}, \bibfield{editor}{\bibinfo{person}{D.~Lee},
  \bibinfo{person}{M.~Sugiyama}, \bibinfo{person}{U.~Luxburg},
  \bibinfo{person}{I.~Guyon}, {and} \bibinfo{person}{R.~Garnett}} (Eds.),
  Vol.~\bibinfo{volume}{29}. \bibinfo{publisher}{Curran Associates, Inc.}
\newblock
\urldef\tempurl%
\url{https://proceedings.neurips.cc/paper/2016/file/55b1927fdafef39c48e5b73b5d61ea60-Paper.pdf}
\showURL{%
\tempurl}


\bibitem[\protect\citeauthoryear{Terry, Grammel, Hari, Santos, Black, and
  Manocha}{Terry et~al\mbox{.}}{2020}]%
        {DBLP:journals/corr/abs-2005-13625}
\bibfield{author}{\bibinfo{person}{Justin~K. Terry}, \bibinfo{person}{Nathaniel
  Grammel}, \bibinfo{person}{Ananth Hari}, \bibinfo{person}{Luis Santos},
  \bibinfo{person}{Benjamin Black}, {and} \bibinfo{person}{Dinesh Manocha}.}
  \bibinfo{year}{2020}\natexlab{}.
\newblock \showarticletitle{Parameter Sharing is Surprisingly Useful for
  Multi-Agent Deep Reinforcement Learning}.
\newblock \bibinfo{journal}{\emph{CoRR}}  \bibinfo{volume}{abs/2005.13625}
  (\bibinfo{year}{2020}).
\newblock
\showeprint[arXiv]{2005.13625}


\bibitem[\protect\citeauthoryear{Tobin, Fong, Ray, Schneider, Zaremba, and
  Abbeel}{Tobin et~al\mbox{.}}{2017a}]%
        {tobin2017domain}
\bibfield{author}{\bibinfo{person}{Josh Tobin}, \bibinfo{person}{Rachel Fong},
  \bibinfo{person}{Alex Ray}, \bibinfo{person}{Jonas Schneider},
  \bibinfo{person}{Wojciech Zaremba}, {and} \bibinfo{person}{Pieter Abbeel}.}
  \bibinfo{year}{2017}\natexlab{a}.
\newblock \showarticletitle{Domain randomization for transferring deep neural
  networks from simulation to the real world}. In
  \bibinfo{booktitle}{\emph{2017 IEEE/RSJ international conference on
  intelligent robots and systems (IROS)}}. IEEE, \bibinfo{pages}{23--30}.
\newblock


\bibitem[\protect\citeauthoryear{Tobin, Fong, Ray, Schneider, Zaremba, and
  Abbeel}{Tobin et~al\mbox{.}}{2017b}]%
        {DBLP:journals/corr/TobinFRSZA17}
\bibfield{author}{\bibinfo{person}{Joshua Tobin}, \bibinfo{person}{Rachel
  Fong}, \bibinfo{person}{Alex Ray}, \bibinfo{person}{Jonas Schneider},
  \bibinfo{person}{Wojciech Zaremba}, {and} \bibinfo{person}{Pieter Abbeel}.}
  \bibinfo{year}{2017}\natexlab{b}.
\newblock \showarticletitle{Domain Randomization for Transferring Deep Neural
  Networks from Simulation to the Real World}.
\newblock \bibinfo{journal}{\emph{CoRR}}  \bibinfo{volume}{abs/1703.06907}
  (\bibinfo{year}{2017}).
\newblock
\showeprint[arXiv]{1703.06907}
\urldef\tempurl%
\url{http://arxiv.org/abs/1703.06907}
\showURL{%
\tempurl}


\bibitem[\protect\citeauthoryear{Travnik, Mathewson, Sutton, and
  Pilarski}{Travnik et~al\mbox{.}}{2018}]%
        {10.3389/frobt.2018.00079}
\bibfield{author}{\bibinfo{person}{Jaden~B. Travnik}, \bibinfo{person}{Kory~W.
  Mathewson}, \bibinfo{person}{Richard~S. Sutton}, {and}
  \bibinfo{person}{Patrick~M. Pilarski}.} \bibinfo{year}{2018}\natexlab{}.
\newblock \showarticletitle{Reactive Reinforcement Learning in Asynchronous
  Environments}.
\newblock \bibinfo{journal}{\emph{Frontiers in Robotics and AI}}
  \bibinfo{volume}{5} (\bibinfo{year}{2018}).
\newblock
\showISSN{2296-9144}
\urldef\tempurl%
\url{https://doi.org/10.3389/frobt.2018.00079}
\showDOI{\tempurl}


\bibitem[\protect\citeauthoryear{Umari and Mukhopadhyay}{Umari and
  Mukhopadhyay}{2017}]%
        {RRT}
\bibfield{author}{\bibinfo{person}{Hassan Umari} {and} \bibinfo{person}{Shayok
  Mukhopadhyay}.} \bibinfo{year}{2017}\natexlab{}.
\newblock \showarticletitle{Autonomous robotic exploration based on multiple
  rapidly-exploring randomized trees}. In \bibinfo{booktitle}{\emph{2017
  IEEE/RSJ International Conference on Intelligent Robots and Systems (IROS)}}.
  \bibinfo{pages}{1396--1402}.
\newblock
\urldef\tempurl%
\url{https://doi.org/10.1109/IROS.2017.8202319}
\showDOI{\tempurl}


\bibitem[\protect\citeauthoryear{Vaswani, Shazeer, Parmar, Uszkoreit, Jones,
  Gomez, Kaiser, and Polosukhin}{Vaswani et~al\mbox{.}}{2017a}]%
        {attention2}
\bibfield{author}{\bibinfo{person}{Ashish Vaswani}, \bibinfo{person}{Noam
  Shazeer}, \bibinfo{person}{Niki Parmar}, \bibinfo{person}{Jakob Uszkoreit},
  \bibinfo{person}{Llion Jones}, \bibinfo{person}{Aidan~N Gomez},
  \bibinfo{person}{{\L}ukasz Kaiser}, {and} \bibinfo{person}{Illia
  Polosukhin}.} \bibinfo{year}{2017}\natexlab{a}.
\newblock \showarticletitle{Attention is all you need}. In
  \bibinfo{booktitle}{\emph{Advances in neural information processing
  systems}}. \bibinfo{pages}{5998--6008}.
\newblock


\bibitem[\protect\citeauthoryear{Vaswani, Shazeer, Parmar, Uszkoreit, Jones,
  Gomez, Kaiser, and Polosukhin}{Vaswani et~al\mbox{.}}{2017b}]%
        {transformer}
\bibfield{author}{\bibinfo{person}{Ashish Vaswani}, \bibinfo{person}{Noam
  Shazeer}, \bibinfo{person}{Niki Parmar}, \bibinfo{person}{Jakob Uszkoreit},
  \bibinfo{person}{Llion Jones}, \bibinfo{person}{Aidan~N Gomez},
  \bibinfo{person}{\L~ukasz Kaiser}, {and} \bibinfo{person}{Illia Polosukhin}.}
  \bibinfo{year}{2017}\natexlab{b}.
\newblock \showarticletitle{Attention is All you Need}. In
  \bibinfo{booktitle}{\emph{Advances in Neural Information Processing
  Systems}}, \bibfield{editor}{\bibinfo{person}{I.~Guyon},
  \bibinfo{person}{U.~V. Luxburg}, \bibinfo{person}{S.~Bengio},
  \bibinfo{person}{H.~Wallach}, \bibinfo{person}{R.~Fergus},
  \bibinfo{person}{S.~Vishwanathan}, {and} \bibinfo{person}{R.~Garnett}}
  (Eds.), Vol.~\bibinfo{volume}{30}. \bibinfo{publisher}{Curran Associates,
  Inc.}
\newblock


\bibitem[\protect\citeauthoryear{von Stumberg, Usenko, Engel, St{\"u}ckler, and
  Cremers}{von Stumberg et~al\mbox{.}}{2017}]%
        {drone}
\bibfield{author}{\bibinfo{person}{Lukas von Stumberg},
  \bibinfo{person}{Vladyslav Usenko}, \bibinfo{person}{Jakob Engel},
  \bibinfo{person}{J{\"o}rg St{\"u}ckler}, {and} \bibinfo{person}{Daniel
  Cremers}.} \bibinfo{year}{2017}\natexlab{}.
\newblock \showarticletitle{From monocular SLAM to autonomous drone
  exploration}. In \bibinfo{booktitle}{\emph{2017 European Conference on Mobile
  Robots (ECMR)}}. IEEE, \bibinfo{pages}{1--8}.
\newblock


\bibitem[\protect\citeauthoryear{Wang, Wang, Zhu, Dai, and Wang}{Wang
  et~al\mbox{.}}{2021}]%
        {multiagent-navigation1}
\bibfield{author}{\bibinfo{person}{Haiyang Wang}, \bibinfo{person}{Wenguan
  Wang}, \bibinfo{person}{Xizhou Zhu}, \bibinfo{person}{Jifeng Dai}, {and}
  \bibinfo{person}{Liwei Wang}.} \bibinfo{year}{2021}\natexlab{}.
\newblock \showarticletitle{Collaborative Visual Navigation}.
\newblock \bibinfo{journal}{\emph{arXiv preprint arXiv:2107.01151}}
  (\bibinfo{year}{2021}).
\newblock


\bibitem[\protect\citeauthoryear{Wang*, Wang*, Wu, and Zhang}{Wang*
  et~al\mbox{.}}{2020}]%
        {Wang*2020Influence-Based}
\bibfield{author}{\bibinfo{person}{Tonghan Wang*}, \bibinfo{person}{Jianhao
  Wang*}, \bibinfo{person}{Yi Wu}, {and} \bibinfo{person}{Chongjie Zhang}.}
  \bibinfo{year}{2020}\natexlab{}.
\newblock \showarticletitle{Influence-Based Multi-Agent Exploration}. In
  \bibinfo{booktitle}{\emph{International Conference on Learning
  Representations}}.
\newblock


\bibitem[\protect\citeauthoryear{Wang, Yang, Liu, Hao, Hao, Hu, Chen, Fan, and
  Gao}{Wang et~al\mbox{.}}{2020a}]%
        {DBLP:conf/aaai/WangYLHHHCFG20}
\bibfield{author}{\bibinfo{person}{Weixun Wang}, \bibinfo{person}{Tianpei
  Yang}, \bibinfo{person}{Yong Liu}, \bibinfo{person}{Jianye Hao},
  \bibinfo{person}{Xiaotian Hao}, \bibinfo{person}{Yujing Hu},
  \bibinfo{person}{Yingfeng Chen}, \bibinfo{person}{Changjie Fan}, {and}
  \bibinfo{person}{Yang Gao}.} \bibinfo{year}{2020}\natexlab{a}.
\newblock \showarticletitle{From Few to More: Large-Scale Dynamic Multiagent
  Curriculum Learning}. In \bibinfo{booktitle}{\emph{The Thirty-Fourth {AAAI}
  Conference on Artificial Intelligence, {AAAI} 2020, The Thirty-Second
  Innovative Applications of Artificial Intelligence Conference, {IAAI} 2020,
  The Tenth {AAAI} Symposium on Educational Advances in Artificial
  Intelligence, {EAAI} 2020, New York, NY, USA, February 7-12, 2020}}.
  \bibinfo{publisher}{{AAAI} Press}, \bibinfo{pages}{7293--7300}.
\newblock


\bibitem[\protect\citeauthoryear{Wang, Yang, Liu, Hao, Hao, Hu, Chen, Fan, and
  Gao}{Wang et~al\mbox{.}}{2020b}]%
        {wang2020few}
\bibfield{author}{\bibinfo{person}{Weixun Wang}, \bibinfo{person}{Tianpei
  Yang}, \bibinfo{person}{Yong Liu}, \bibinfo{person}{Jianye Hao},
  \bibinfo{person}{Xiaotian Hao}, \bibinfo{person}{Yujing Hu},
  \bibinfo{person}{Yingfeng Chen}, \bibinfo{person}{Changjie Fan}, {and}
  \bibinfo{person}{Yang Gao}.} \bibinfo{year}{2020}\natexlab{b}.
\newblock \showarticletitle{From Few to More: Large-Scale Dynamic Multiagent
  Curriculum Learning}. In \bibinfo{booktitle}{\emph{Proceedings of the AAAI
  Conference on Artificial Intelligence}}, Vol.~\bibinfo{volume}{34}.
  \bibinfo{pages}{7293--7300}.
\newblock


\bibitem[\protect\citeauthoryear{Wang, Girshick, Gupta, and He}{Wang
  et~al\mbox{.}}{2018}]%
        {wang2018non}
\bibfield{author}{\bibinfo{person}{Xiaolong Wang}, \bibinfo{person}{Ross
  Girshick}, \bibinfo{person}{Abhinav Gupta}, {and} \bibinfo{person}{Kaiming
  He}.} \bibinfo{year}{2018}\natexlab{}.
\newblock \showarticletitle{Non-local neural networks}. In
  \bibinfo{booktitle}{\emph{Proceedings of the IEEE conference on computer
  vision and pattern recognition}}. \bibinfo{pages}{7794--7803}.
\newblock


\bibitem[\protect\citeauthoryear{Wurm, Stachniss, and Burgard}{Wurm
  et~al\mbox{.}}{2008}]%
        {multi-classical1}
\bibfield{author}{\bibinfo{person}{Kai~M Wurm}, \bibinfo{person}{Cyrill
  Stachniss}, {and} \bibinfo{person}{Wolfram Burgard}.}
  \bibinfo{year}{2008}\natexlab{}.
\newblock \showarticletitle{Coordinated multi-robot exploration using a
  segmentation of the environment}. In \bibinfo{booktitle}{\emph{2008 IEEE/RSJ
  International Conference on Intelligent Robots and Systems}}. IEEE,
  \bibinfo{pages}{1160--1165}.
\newblock


\bibitem[\protect\citeauthoryear{Xia, Zamir, He, Sax, Malik, and Savarese}{Xia
  et~al\mbox{.}}{2018}]%
        {dataset}
\bibfield{author}{\bibinfo{person}{Fei Xia}, \bibinfo{person}{Amir~R Zamir},
  \bibinfo{person}{Zhiyang He}, \bibinfo{person}{Alexander Sax},
  \bibinfo{person}{Jitendra Malik}, {and} \bibinfo{person}{Silvio Savarese}.}
  \bibinfo{year}{2018}\natexlab{}.
\newblock \showarticletitle{Gibson env: Real-world perception for embodied
  agents}. In \bibinfo{booktitle}{\emph{Proceedings of the IEEE Conference on
  Computer Vision and Pattern Recognition}}. \bibinfo{pages}{9068--9079}.
\newblock


\bibitem[\protect\citeauthoryear{Xiao, Tan, and Amato}{Xiao
  et~al\mbox{.}}{2022}]%
        {https://doi.org/10.48550/arxiv.2209.10113}
\bibfield{author}{\bibinfo{person}{Yuchen Xiao}, \bibinfo{person}{Weihao Tan},
  {and} \bibinfo{person}{Christopher Amato}.} \bibinfo{year}{2022}\natexlab{}.
\newblock \bibinfo{title}{Asynchronous Actor-Critic for Multi-Agent
  Reinforcement Learning}.
\newblock
\newblock
\urldef\tempurl%
\url{https://doi.org/10.48550/ARXIV.2209.10113}
\showDOI{\tempurl}


\bibitem[\protect\citeauthoryear{Yamauchi}{Yamauchi}{1997a}]%
        {yamauchi1997frontier}
\bibfield{author}{\bibinfo{person}{Brian Yamauchi}.}
  \bibinfo{year}{1997}\natexlab{a}.
\newblock \showarticletitle{A frontier-based approach for autonomous
  exploration}. In \bibinfo{booktitle}{\emph{Proceedings 1997 IEEE
  International Symposium on Computational Intelligence in Robotics and
  Automation CIRA'97.'Towards New Computational Principles for Robotics and
  Automation'}}. IEEE, \bibinfo{pages}{146--151}.
\newblock


\bibitem[\protect\citeauthoryear{Yamauchi}{Yamauchi}{1997b}]%
        {frontier2}
\bibfield{author}{\bibinfo{person}{Brian Yamauchi}.}
  \bibinfo{year}{1997}\natexlab{b}.
\newblock \showarticletitle{A frontier-based approach for autonomous
  exploration}. In \bibinfo{booktitle}{\emph{Proceedings 1997 IEEE
  International Symposium on Computational Intelligence in Robotics and
  Automation CIRA'97.'Towards New Computational Principles for Robotics and
  Automation'}}. IEEE, \bibinfo{pages}{146--151}.
\newblock


\bibitem[\protect\citeauthoryear{Yang, Nakhaei, Isele, Fujimura, and Zha}{Yang
  et~al\mbox{.}}{2020}]%
        {Yang2020CM3}
\bibfield{author}{\bibinfo{person}{Jiachen Yang}, \bibinfo{person}{Alireza
  Nakhaei}, \bibinfo{person}{David Isele}, \bibinfo{person}{Kikuo Fujimura},
  {and} \bibinfo{person}{Hongyuan Zha}.} \bibinfo{year}{2020}\natexlab{}.
\newblock \showarticletitle{CM3: Cooperative Multi-goal Multi-stage Multi-agent
  Reinforcement Learning}. In \bibinfo{booktitle}{\emph{International
  Conference on Learning Representations}}.
\newblock


\bibitem[\protect\citeauthoryear{Yao, Yin, Yang, Yu, Shen, Zhang, Liang, and
  Huang}{Yao et~al\mbox{.}}{2021}]%
        {yao2021partially}
\bibfield{author}{\bibinfo{person}{Meng Yao}, \bibinfo{person}{Qiyue Yin},
  \bibinfo{person}{Jun Yang}, \bibinfo{person}{Tongtong Yu},
  \bibinfo{person}{Shengqi Shen}, \bibinfo{person}{Junge Zhang},
  \bibinfo{person}{Bin Liang}, {and} \bibinfo{person}{Kaiqi Huang}.}
  \bibinfo{year}{2021}\natexlab{}.
\newblock \showarticletitle{The Partially Observable Asynchronous Multi-Agent
  Cooperation Challenge}.
\newblock \bibinfo{journal}{\emph{arXiv preprint arXiv:2112.03809}}
  (\bibinfo{year}{2021}).
\newblock


\bibitem[\protect\citeauthoryear{Yu, Velu, Vinitsky, Wang, Bayen, and Wu}{Yu
  et~al\mbox{.}}{2021b}]%
        {mappo}
\bibfield{author}{\bibinfo{person}{Chao Yu}, \bibinfo{person}{Akash Velu},
  \bibinfo{person}{Eugene Vinitsky}, \bibinfo{person}{Yu Wang},
  \bibinfo{person}{Alexandre Bayen}, {and} \bibinfo{person}{Yi Wu}.}
  \bibinfo{year}{2021}\natexlab{b}.
\newblock \showarticletitle{The surprising effectiveness of mappo in
  cooperative, multi-agent games}.
\newblock \bibinfo{journal}{\emph{arXiv preprint arXiv:2103.01955}}
  (\bibinfo{year}{2021}).
\newblock


\bibitem[\protect\citeauthoryear{Yu, Velu, Vinitsky, Wang, Bayen, and Wu}{Yu
  et~al\mbox{.}}{2021c}]%
        {surprising2021yu}
\bibfield{author}{\bibinfo{person}{Chao Yu}, \bibinfo{person}{Akash Velu},
  \bibinfo{person}{Eugene Vinitsky}, \bibinfo{person}{Yu Wang},
  \bibinfo{person}{Alexandre Bayen}, {and} \bibinfo{person}{Yi Wu}.}
  \bibinfo{year}{2021}\natexlab{c}.
\newblock \bibinfo{title}{The Surprising Effectiveness of PPO in Cooperative,
  Multi-Agent Games}.
\newblock
\newblock
\urldef\tempurl%
\url{https://doi.org/10.48550/ARXIV.2103.01955}
\showDOI{\tempurl}


\bibitem[\protect\citeauthoryear{Yu, Yang, Gao, Yang, Wang, and Wu}{Yu
  et~al\mbox{.}}{2021d}]%
        {maans}
\bibfield{author}{\bibinfo{person}{Chao Yu}, \bibinfo{person}{Xinyi Yang},
  \bibinfo{person}{Jiaxuan Gao}, \bibinfo{person}{Huazhong Yang},
  \bibinfo{person}{Yu Wang}, {and} \bibinfo{person}{Yi Wu}.}
  \bibinfo{year}{2021}\natexlab{d}.
\newblock \showarticletitle{Learning Efficient Multi-Agent Cooperative Visual
  Exploration}.
\newblock \bibinfo{journal}{\emph{arXiv preprint arXiv:2110.05734}}
  (\bibinfo{year}{2021}).
\newblock


\bibitem[\protect\citeauthoryear{Yu, Tong, Xu, Xu, Dong, Yang, and Wang}{Yu
  et~al\mbox{.}}{2021a}]%
        {APF}
\bibfield{author}{\bibinfo{person}{Jincheng Yu}, \bibinfo{person}{Jianming
  Tong}, \bibinfo{person}{Yuanfan Xu}, \bibinfo{person}{Zhilin Xu},
  \bibinfo{person}{Haolin Dong}, \bibinfo{person}{Tianxiang Yang}, {and}
  \bibinfo{person}{Yu Wang}.} \bibinfo{year}{2021}\natexlab{a}.
\newblock \showarticletitle{SMMR-Explore: SubMap-based Multi-Robot Exploration
  System with Multi-robot Multi-target Potential Field Exploration Method}. In
  \bibinfo{booktitle}{\emph{2021 IEEE International Conference on Robotics and
  Automation (ICRA)}}.
\newblock


\bibitem[\protect\citeauthoryear{Zambaldi, Raposo, Santoro, Bapst, Li,
  Babuschkin, Tuyls, Reichert, Lillicrap, Lockhart, et~al\mbox{.}}{Zambaldi
  et~al\mbox{.}}{2018}]%
        {zambaldi2018relational}
\bibfield{author}{\bibinfo{person}{Vinicius Zambaldi}, \bibinfo{person}{David
  Raposo}, \bibinfo{person}{Adam Santoro}, \bibinfo{person}{Victor Bapst},
  \bibinfo{person}{Yujia Li}, \bibinfo{person}{Igor Babuschkin},
  \bibinfo{person}{Karl Tuyls}, \bibinfo{person}{David Reichert},
  \bibinfo{person}{Timothy Lillicrap}, \bibinfo{person}{Edward Lockhart},
  {et~al\mbox{.}}} \bibinfo{year}{2018}\natexlab{}.
\newblock \showarticletitle{Relational deep reinforcement learning}.
\newblock \bibinfo{journal}{\emph{arXiv preprint arXiv:1806.01830}}
  (\bibinfo{year}{2018}).
\newblock


\bibitem[\protect\citeauthoryear{Zhang, Song, Huang, Swami, and Chawla}{Zhang
  et~al\mbox{.}}{2019b}]%
        {graphnetwork}
\bibfield{author}{\bibinfo{person}{Chuxu Zhang}, \bibinfo{person}{Dongjin
  Song}, \bibinfo{person}{Chao Huang}, \bibinfo{person}{Ananthram Swami}, {and}
  \bibinfo{person}{Nitesh~V Chawla}.} \bibinfo{year}{2019}\natexlab{b}.
\newblock \showarticletitle{Heterogeneous graph neural network}. In
  \bibinfo{booktitle}{\emph{Proceedings of the 25th ACM SIGKDD International
  Conference on Knowledge Discovery \& Data Mining}}.
  \bibinfo{pages}{793--803}.
\newblock


\bibitem[\protect\citeauthoryear{Zhang, Yang, and Ba{\c{s}}ar}{Zhang
  et~al\mbox{.}}{2021}]%
        {zhang2021multi}
\bibfield{author}{\bibinfo{person}{Kaiqing Zhang}, \bibinfo{person}{Zhuoran
  Yang}, {and} \bibinfo{person}{Tamer Ba{\c{s}}ar}.}
  \bibinfo{year}{2021}\natexlab{}.
\newblock \showarticletitle{Multi-agent reinforcement learning: A selective
  overview of theories and algorithms}.
\newblock \bibinfo{journal}{\emph{Handbook of Reinforcement Learning and
  Control}} (\bibinfo{year}{2021}), \bibinfo{pages}{321--384}.
\newblock


\bibitem[\protect\citeauthoryear{Zhang, Hare, and Prugel-Bennett}{Zhang
  et~al\mbox{.}}{2019a}]%
        {deepset}
\bibfield{author}{\bibinfo{person}{Yan Zhang}, \bibinfo{person}{Jonathon Hare},
  {and} \bibinfo{person}{Adam Prugel-Bennett}.}
  \bibinfo{year}{2019}\natexlab{a}.
\newblock \showarticletitle{Deep set prediction networks}.
\newblock \bibinfo{journal}{\emph{Advances in Neural Information Processing
  Systems}}  \bibinfo{volume}{32} (\bibinfo{year}{2019}),
  \bibinfo{pages}{3212--3222}.
\newblock


\bibitem[\protect\citeauthoryear{Zhu, Hu, Zhang, Hong, Zhu, Chang, and
  Liang}{Zhu et~al\mbox{.}}{2021}]%
        {multiagent-RL2}
\bibfield{author}{\bibinfo{person}{Fengda Zhu}, \bibinfo{person}{Siyi Hu},
  \bibinfo{person}{Yi Zhang}, \bibinfo{person}{Haodong Hong},
  \bibinfo{person}{Yi Zhu}, \bibinfo{person}{Xiaojun Chang}, {and}
  \bibinfo{person}{Xiaodan Liang}.} \bibinfo{year}{2021}\natexlab{}.
\newblock \showarticletitle{MAIN: A Multi-agent Indoor Navigation Benchmark for
  Cooperative Learning}.
\newblock  (\bibinfo{year}{2021}).
\newblock


\end{thebibliography}


\end{document}